\documentclass[journal]{IEEEtran}
\newcommand{\ignore}[1]{}
\usepackage[utf8]{inputenc}
\usepackage[T1]{fontenc}
\usepackage{graphicx}
\usepackage{amsmath}
\usepackage{amssymb}
\usepackage{booktabs}
\usepackage{times}
\usepackage{epsfig}
\usepackage{cite}

\usepackage{algorithm}  
\usepackage{algpseudocode}
\usepackage{xcolor}
\usepackage{pifont}
\usepackage{multirow}
\usepackage{bbm}
\usepackage{mathtools}
\newcommand{\cmark}{\ding{51}}

\newcommand{\etal}{\textit{et al}. }

\DeclareMathOperator*{\argmax}{arg\,max}

\usepackage[pagebackref=true,breaklinks=true,colorlinks,bookmarks=false,linkcolor=black,anchorcolor=black,citecolor=black]{hyperref}

\begin{document}
\title{Deep Boosting Learning: A Brand-new Cooperative Approach for Image-Text Matching}
\author{Haiwen~Diao,
        Ying~Zhang,
        Shang~Gao,
        Xiang~Ruan,
        Huchuan~Lu
\thanks{
\textit{Corresponding author: Huchuan Lu (lhchuan@dlut.edu.cn).} 
This work was supported by the National Natural Science Foundation of China under grant No. 62293540, 62293542.
H. Diao, S. Gao are with Dalian University of Technology, China. (Email: diaohw@mail.dlut.edu.cn; gs940601k@gmail.com).
Y. Zhang is with Tencent Company, China.
(Email: yinggzhang@tencent.com).
X. Ruan is with Tiwaki Company, Japan.
(Email: ruanxiang@tiwaki.com).
}}
\maketitle
\markboth{IEEE Transactions on Image Processing}{}

\begin{abstract}
Image-text matching remains a challenging task due to heterogeneous semantic diversity across modalities and insufficient distance separability within triplets. Different from previous approaches focusing on enhancing multi-modal representations or exploiting cross-modal correspondence for more accurate retrieval, in this paper we aim to leverage the knowledge transfer between peer branches in a boosting manner to seek a more powerful matching model. Specifically, we propose a brand-new Deep Boosting Learning (DBL) algorithm, where an anchor branch is first trained to provide insights into the data properties, with a target branch gaining more advanced knowledge to develop optimal features and distance metrics.
Concretely, an anchor branch initially learns the absolute or relative distance between positive and negative pairs, providing a foundational understanding of the particular network and data distribution. Building upon this knowledge, a target branch is concurrently tasked with more adaptive margin constraints to further enlarge the relative distance between matched and unmatched samples.
Extensive experiments validate that our DBL can achieve impressive and consistent improvements based on various recent state-of-the-art models in the image-text matching field, and outperform related popular cooperative strategies, e.g., Conventional Distillation, Mutual Learning, and Contrastive Learning. Beyond the above, we confirm that DBL can be seamlessly integrated into their training scenarios and achieve superior performance under the same computational costs, demonstrating the flexibility and broad applicability of our proposed method.
\end{abstract}
\IEEEpeerreviewmaketitle


\section{Introduction}
With the explosion of multimedia volume in recent years, image-text matching~\cite{ITM:DeViSE,ITM:UVSE} has been a prevalent research topic, which efficiently bridges the gap between vision and language, and potentially benefits other multi-modal tasks such as video-text retrieval~\cite{VLP:HGR,VLP:EHLS,VLP:HIT}, referring expression~\cite{REC:LGGAN,REC:CKR}, and visual question answering~\cite{VQA:MCAN,VQA:VQA-A}, etc. Despite years of efforts, image-text matching remains challenging because it entails not only recognizing hierarchical contents across modalities~\cite{ITM:MMCA,ITM:CAAN}, but also mapping diverse inputs into a comparable space to exploit semantic associations~\cite{ITM:SCO,ITM:GXN,ITM:AME}.

To explore a shared embedding space for cross-modal data, some works~\cite{ITM:DSPE,VQA:DAN} employ a hinge-based ranking loss function which forces each image/text to be closer to its positive text/image than all negatives within a mini-batch. Each triplet would be punished when the relative distance between query-positive and query-negative pairs is less than a fixed margin.
Though considering all pairs makes the optimization more stable, its sum-margin strategy treats all triplets equally during optimization, which would diminish the impact of valuable ones.
To excavate more informative matching details, Faghri \etal~\cite{ITM:VSE++} and Wei \etal~\cite{ITM:MPL} propose the max-margin and polynomial loss respectively to assign appropriate weights and highlight significant pairs from redundant pairs. However, the fixed distance margin for all triplets does not necessarily lead to good separability between the positive and negative samples. Hence, Zhao \etal~\cite{ML:ATL} employs adaptive thresholds by computing the feature distances between text-to-text pairs as a reference, while Biten \etal~\cite{ITM:SAM} takes captioning metric (SPICE or CIDEr) as a measure and generates the semantic boundary via the language continuum of each caption. Besides, Zhou \etal~\cite{ITM:Ladder} proposes the ladder loss with an inequality chain and adopts hierarchical margins for all triplets.
However, they all consider an implicit and coarse relevance representation between each query and its candidates, resulting in an imprecise and inconsistent threshold constraint, and matching ambiguities between positive and negative pairs.

From the above perspective, we reconsider the key ingredients to fully exploit the potential of a matching network, i.e., constructing specialized guidance and seeking appropriate penalties. In this paper, we propose a novel Deep Boosting Learning (DBL) strategy, where an anchor branch is trained synchronously or asynchronously to provide explicit adaptive constraint for each triplet, in order to obtain a more powerful target branch. Different from previous metrics imposing inexplicit handcrafted penalties, the anchor branch learns the distance distribution and triplet relationship from data in advance, and would naturally gain an insight into the model properties and matching patterns, offering its twin target branch an explicit measurement to further enlarge distance separation and gain more discriminative feature metric. More specifically, the penalty of each triplet would be adjusted dynamically according to the similarity values predicted by the anchor branch, aiming to increase the association/separability between matched/unmatched pairs, as well as relax the tedious and costly configurations for robust constraint exploration. In this way, the target branch would capture a comprehensive picture of data and model characteristics, and translate the prior distance reference into more powerful matching capacities.

We notice that the proposed DBL is closely relevant to previous cooperative learning strategies including Conventional Distillation, Mutual Learning, and Contrastive Learning - they attempt to transfer prior knowledge and achieve better performance across multiple branches compared with independent learning. Specifically, they typically adopt a static pre-trained network~\cite{KD:Fitnet,KD:AT,KD:FSP}, peer-teaching cohorts~\cite{KD:DML}, or a slowly progressing encoder~\cite{CL:MoCo,CL:BYOL,CL:DINO} as the anchor branch. To learn better probability prediction or feature representation, the target branch is encouraged to mimic the outputs of the anchor branch, including hard/soft pseudo-labels~\cite{KD:KD,KD:Darkrank,KD:DML}, absolute/relative relations~\cite{KD:SP,KD:RKD,KD:LMFT}, and feature similarities~\cite{CL:SimCLR,CL:SimSiam,CL:BYOL}. However, these approaches are originally designed for uni-modal tasks, and none of them take into account the heterogeneous semantic gap in cross-modal data. 
Besides, our DBL goes deeper into the margin knowledge in a peer-boosting manner beyond their peer-imitating ways to gain greater benefits under the same training and inference schemes, confirming the necessity of investigating effective cooperative strategies for cross-modal retrieval.

Our contributions are summarized as follows:
\begin{itemize}
    \item We propose a novel Deep Boosting Learning (DBL) for peer-training strategy, which introduces an adaptive and explicit margin constraint for each triplet, and effectively generates the initiative distance separability between positive and negative pairs for image-text matching.
    \item Our DBL strategy can be widely applied to multiple training scenarios of related cooperative approaches, either as a post-processing step or in an online manner via collaborative or momentum synchronous updates.
    \item We validate the proposed DBL strategy with recent state-of-the-art works. Extensive experimental results on Flickr30K and MSCOCO datasets demonstrate the superiority and flexibility of our boosting strategy.
\end{itemize}

\section{Related work}
\subsection{Image-Text Matching.} 
Image-text matching task targets retrieving images from the database with natural language queries, and vice versa. Research on this topic can be roughly divided into two aspects: \textbf{1) mono-modal representation.} To achieve this, some works~\cite{ITM:VSRN,ITM:HOAD,ITM:WCGL,ITM:HAT} introduced graph reasoning networks to enhance the region and word features, while Wang \etal~\cite{ITM:CODER} utilized a constructed concept correlation to generate the consensus-aware embeddings. Besides, Li \etal~\cite{ITM:MEMBER}, Chen \etal~\cite{ITM:GPO} and Zhu \etal~\cite{ITM:ESA} designed global memory bank, generalized pooling operator, and external space attention strategy respectively which effectively enhance the feature representation and facilitate mono-modal aggregation. Another set of 
methods~\cite{ITM:SCAN,ITM:BFAN,ITM:PFAN,ITM:DIME} focuses on \textbf{2) cross-modal interaction}. For example, some approaches~\cite{ITM:IMRAM,ITM:RCAR} employed an iterative scheme with attention memory or regulator modules to recurrently refine region-word alignments, while several methods~\cite{ITM:GSMN,ITM:SGRAF,ITM:CMCAN,ITM:HAT} developed cross-modal correspondences to perform hierarchical matching with complex graph reasoning and high computational cost. Moreover, Zhang \etal~\cite{ITM:NAAF} used the optimal boundary to explicitly and adaptively model the mismatched fragments and yield more accurate predictions. To evaluate the effectiveness and generalization of our proposed strategy, we apply our DBL to a series of representative works including conventional and pre-trained matching networks on the above two directions, and achieve solid and consistent improvements on two benchmarks. Note that we also construct a concise but powerful baseline, which, though not our contribution, only serves as an insight into the mechanisms and comparisons of our DBL strategy.

\subsection{Deep Metric Learning.}
Deep metric learning aims to map samples into a unified projection space, such that the similarities between positive pairs are higher than the ones between negative pairs. In the past few years, various loss functions for uni-modal retrieval tasks~\cite{ReID:HardTriplet,ReID:Quadruplet} have been introduced, including triplet~\cite{ReID:HardTriplet}, quadruplet~\cite{ReID:Quadruplet}, lifted structure~\cite{ML:Lifted}, N-pair~\cite{ML:N-pair}, histogram~\cite{ML:Histogram}, and Proxy-NCA~\cite{ML:Proxy-NCA}, some of which have been extended for cross-modal matching. Wang \etal~\cite{ITM:DSPE} proposed a two-way ranking loss by adapting the triplet loss for bi-directional retrievals, which has gained great popularity in multi-modal learning~\cite{LVM:MCN,TPS:GNA-RNN}.
Faghri \etal~\cite{ITM:VSE++} introduced hard negative mining into the loss function, while Zhang \etal~\cite{ITM:CMPL} developed the cross-modal projection loss, which minimizes the matching distribution between all pairs in a mini-batch. As mentioned before, the most related works are \cite{ITM:SAM,ITM:Ladder}, which regarded the relationship between text-to-text pairs as a reference and employed adaptive margin restrictions based on implicit semantic relevance degrees. Different from them, the DBL strategy automatically seeks an appropriate threshold according to the explicit distance within a triplet, and forces an adequate distance separation between matched and unmatched image-text pairs.

\subsection{Deep Cooperative Learning.}
Deep cooperative learning is a typical peer-training strategy trading training efficiency for performance benefits, which brings extra training costs but no computational cost for inference. Specifically, conventional distillation trains a smaller student network to mimic the knowledge flow of a powerful yet static teacher, consisting of normalized probabilities~\cite{KD:KD,KD:Darkrank}, network parameters~\cite{KD:Fitnet,KD:AT,KD:KDSVD}, and feature relations~\cite{KD:SP,KD:RKD}, while in mutual learning~\cite{KD:DML}, the student cohorts imitate the predictions from each other, and their training processes are collaborative. Moreover, contrastive learning~\cite{CL:SimCLR,CL:MoCo,CL:BYOL,CL:DINO} presents a promising way of unsupervised representation learning, and the key idea is constructing similar or dissimilar data examples and maximizing agreement between two encoder networks. In contrast, our DBL strategy starts with an anchor branch, first snooping on the prior relationships within triplets, and punishes a target branch via adaptive and adequate margin values. By this means, the latter can obtain a greater discriminative ability and achieve a better matching capability for single-branch learning. Besides, we experimentally validate that it can perform particularly well under multiple training scenarios of the above-mentioned approaches, reflecting the flexibility and wide suitability of our DBL strategy.

\begin{figure}[t!]
	\centering
	\begin{tabular}{@{}c}
		\includegraphics[width=0.98\linewidth, height=0.55\linewidth,trim= 0 345 600 0,clip]{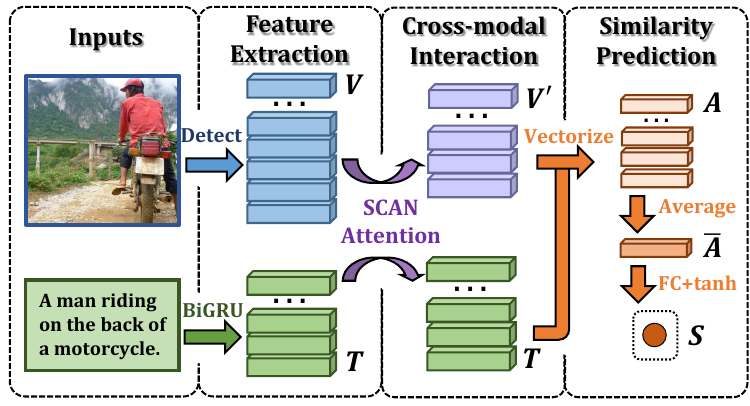}
	\end{tabular}
	\caption{Illustration of our single branch baseline. We adopt hard ranking loss [14] as a task-specific loss to supervise the training process.}
	\label{fig:SBL}
\end{figure}

\section{Methodology}

We first introduce a powerful single-branch network in detail, including feature extraction, cross-modal interaction, similarity prediction, and matching loss. 
We then elaborate on the detailed mathematics and training strategies of our DBL.

\subsection{Single Branch Baseline}
We build a simple and effective image-text matching baseline, which simply combines cross-attention module~\cite{ITM:SCAN} and vectorized similarity representation~\cite{ITM:SGRAF}, and is only used to analyze the mechanism and comparison of our DBL. 

\textbf{Feature Extraction.} For each image, we first apply button-up attention~\cite{IC:BU_TDA} pretrained on Visual Genomes~\cite{Datasets:VisualGenome} to extract the top $K$ region proposals with 2048-d features. Then, a fully-connected (FC) layer is utilized to map these features into 1024-d vectors $\boldsymbol{V}=\{\boldsymbol{v}_{1},...,\boldsymbol{v}_{K}\}\in\mathbbm{R}^{K \times 1024}$. For a sentence with L words, we first encode them into 300-d word embeddings with random initialization, followed by a Bi-GRU~\cite{Baseline:Bi-GRU} to integrate the bidirectional contextual information. Finally, we average the forward and backward hidden states at each time to get the word features $\boldsymbol{T}=\{\boldsymbol{t}_{1},...,\boldsymbol{t}_{L}\}\in\mathbbm{R}^{L \times 1024}$, and $\boldsymbol{t}_{l}$ denotes the $l$-th word vector.

\textbf{Cross-modal Interaction.} We employ widely-used cross-modal attention~\cite{ITM:SCAN} to capture region-word correspondence. Here, we take text-to-image attention as the backbone. To be specific, we first compute the cosine similarity matrix $\boldsymbol{M}\in\mathbbm{R}^{L \times K}$ for all region-word pairs, followed by a zero threshold and word-wise L2 normalization (norm). Then, we adopt the region-wise softmax function and integrate all the regions attended by each word as:
\begin{equation}
\label{eq:croint}
\begin{split}
\boldsymbol{M} &= norm_{\boldsymbol{T}}(\left [ \boldsymbol{T}\boldsymbol{V}^{\top} \right ]_{+}), \\
\boldsymbol{V}^{\prime} &= softmax_{\boldsymbol{V}}(\lambda \boldsymbol{M})\boldsymbol{V}, 
\end{split}
\end{equation}
where $\lambda$ = $9$ following~\cite{ITM:SCAN}, and $\left [ x \right ]_{+}$ = $max(x, 0)$. Note that $\boldsymbol{V}^{\prime}=\{\boldsymbol{v}^{\prime}_{1},...,\boldsymbol{v}^{\prime}_{L}\} \in\mathbbm{R}^{L \times 1024}$, and $\boldsymbol{v}^{\prime}_{l}$ denotes the attended region features with respect to $l$-th word feature.

\textbf{Similarity Prediction.} As with~\cite{ITM:SGRAF}, we first vectorize all the word-based alignments $\boldsymbol{A}\in\mathbbm{R}^{L \times 256}$ between $\boldsymbol{T}$ and $\boldsymbol{V}^{\prime}$, followed by the average operation to obtain one holistic alignment vector. Finally, we feed it into another FC layer and Tanh activation to output a scalar score:
\begin{equation}
\label{eq:simpre}
\begin{split}
    \boldsymbol{A} &= norm(\boldsymbol{W}_{1}(|\boldsymbol{T}-\boldsymbol{V}^{\prime}|^{2})+\boldsymbol{b}_{1}) \ , \\
    \quad
    \mathcal{S} &= tanh(\boldsymbol{W}_{2}(\bar{\boldsymbol{A}})+\boldsymbol{b}_{2}) \ , 
\end{split} 
\end{equation}
where $| \cdot |^2$ denotes the element-wise square. $\boldsymbol{W}_{\{\cdot\}}$ and $\boldsymbol{b}_{\{\cdot\}}$ are learnable parameters, and $\bar{\boldsymbol{A}}\in\mathbbm{R}^{1 \times 256}$ represents word-wise average of $\boldsymbol{A}\in\mathbbm{R}^{L \times 256}$, which indicates the similarity features attended by sentence words.

\textbf{Matching Loss.} Given a batch $\mathcal{D}=\{(\boldsymbol{i}_{n},\boldsymbol{c}_{n})\}_{n=1}^{N}$ with $N$ image-text pairs, the similarity outputs are denoted as $\mathcal{S}_{\{\cdot,\cdot\}}$. Note that $\mathcal{S}_{\boldsymbol{i},\boldsymbol{c}}$ and $\mathcal{S}_{\grave{\boldsymbol{i}}, \boldsymbol{c}}/\mathcal{S}_{\boldsymbol{i}, \grave{\boldsymbol{c}}}$ represent the matching scores of positive and negative pairs. Then, a hinge-based triplet ranking loss~\cite{VQA:DAN,ITM:VSE++} is widely used to guide optimization as:
\begin{equation}
\mathcal{L}_{raw}=\sum\nolimits_{n=1}^{N}\ell(\boldsymbol{i}_{n},\boldsymbol{c}_{n}).
\end{equation}
\textbf{1) Sum-margin strategy}. It takes into account all possible combinations, ideally forcing all positive and negative samples to be separated by a margin value $\gamma$:
\begin{equation}
\label{eq:sumlos}
\ell_{sum}= \sum\nolimits_{\grave{\boldsymbol{c}}} 
[ \gamma  +  \mathcal{S}_{\boldsymbol{i}, \grave{\boldsymbol{c}}} - \mathcal{S}_{\boldsymbol{i}, \boldsymbol{c}}]_{+}
+ \sum\nolimits_{\grave{\boldsymbol{i}}} [ \gamma + \mathcal{S}_{\grave{\boldsymbol{i}}, \boldsymbol{c}} - \mathcal{S}_{\boldsymbol{i}, \boldsymbol{c}}]_{+},
\end{equation}
where $\grave{\boldsymbol{c}}$ and $\grave{\boldsymbol{i}}$ are the negatives of $\boldsymbol{i}$ and $\boldsymbol{c}$, and $\gamma=0.2$. 

\noindent\textbf{2) Max-margin strategy}. In contrast, the hard form only focuses on the nearest negatives ($\hat{\boldsymbol{i}},\hat{\boldsymbol{c}}$) in a mini-batch $\mathcal{D}$:
\begin{equation}
\label{eq:maxlos}
\ell_{max}=
[ \gamma  +  \mathcal{S}_{\boldsymbol{i}, \hat{\boldsymbol{c}}} - \mathcal{S}_{\boldsymbol{i}, \boldsymbol{c}}]_{+}
+ [ \gamma + \mathcal{S}_{\hat{\boldsymbol{i}}, \boldsymbol{c}} - \mathcal{S}_{\boldsymbol{i}, \boldsymbol{c}}]_{+},
\end{equation}
where $\hat{\boldsymbol{c}}={\argmax}_{d\neq c}\mathcal{S}_{\boldsymbol{i}, \boldsymbol{d}}$, $\hat{\boldsymbol{i}}={\argmax}_{j\neq i}\mathcal{S}_{\boldsymbol{j}, \boldsymbol{c}}$.

\textbf{Discussion.} 
Although the latter can explore more informative details and effectively distinguish the confusing samples than the former, they both employ a handcrafted fixed threshold to restrict the relative distances between positive and negative pairs, resulting in an inadequate regularization that is easy for simple samples and hard for confusing ones. We empirically find that the network capability remains under-explored with such independent single-branch training. 
Hence, we propose the DBL strategy which carefully leverages peer knowledge to achieve greater matching capabilities.

\subsection{Deep Boosting Learning}
The core idea is that, given the absolute or relative distance within triplets of the anchor branch, absolute or relative boosting strategies impose more compelling restraints on the corresponding distance for the target branch respectively, so that the target branch can obtain more suitable margin penalties and learn superior matching patterns not available in the anchor branch. To distinguish two branches, we denote the similarity scores of the target and anchor branch as $\mathcal{S}_{\{\cdot,\cdot\}}^{t}$ and $\mathcal{S}_{\{\cdot,\cdot\}}^{a}$. The cumulative loss of boosting learning over training data $\mathcal{D}$ is defined as:
\begin{equation}
    \mathcal{L}_{boo}=\sum\nolimits_{n=1}^{N}\ell^{\prime}(\boldsymbol{i}_{n},\boldsymbol{c}_{n}) .
\end{equation}

\begin{figure*}[t!]
	\centering
	\begin{tabular}{@{}c}
		\includegraphics[width=0.98\linewidth, height=0.28\linewidth,trim= 0 310 185 0,clip]{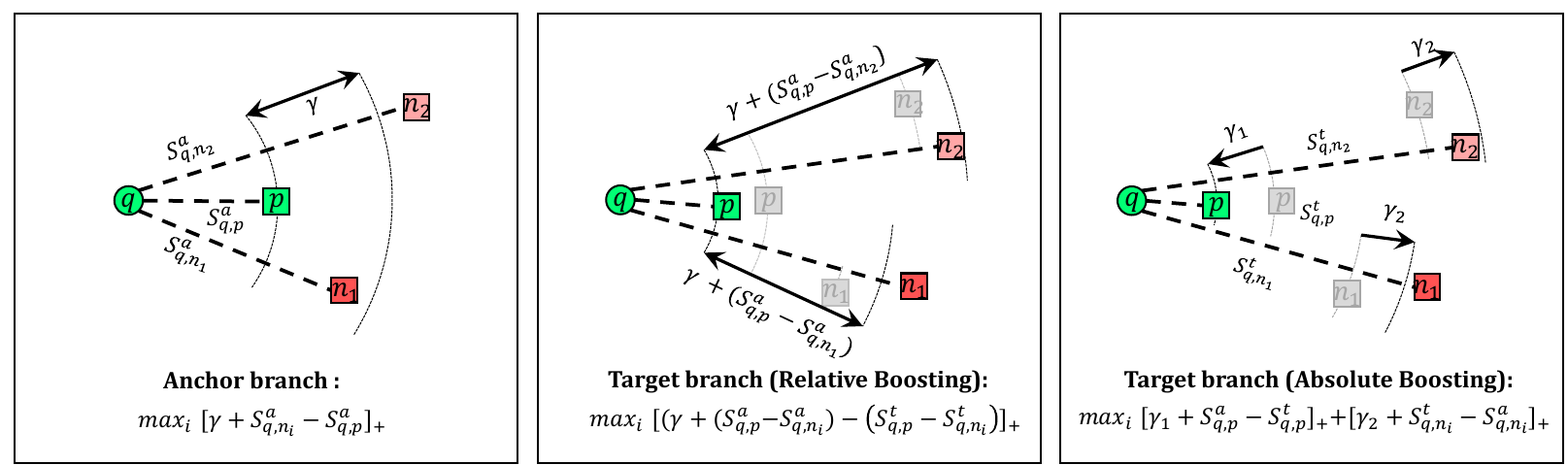} 
	\end{tabular}
	\caption{Illustration of deep boosting learning. We first perform the anchor branch to obtain the absolute distance ($\mathcal{S}_{\boldsymbol{q}, \boldsymbol{p}}^{a},\mathcal{S}_{\boldsymbol{q}, \boldsymbol{n_{i}}}^{a}|_{i=1,2}$) between query and each candidate, and relative distance ($\mathcal{S}_{\boldsymbol{q}, \boldsymbol{p}}^{a} - \mathcal{S}_{\boldsymbol{q}, {\boldsymbol{n_{i}}}}^{a}|_{i=1,2}$) within each triplet. Based on this prior knowledge, we assign the target branch with appropriate thresholds to further enlarge the variations between matched and unmatched image-text pairs.}
	\label{fig:DBL}
\end{figure*}

\textbf{Relative Boosting Strategy.} To fully make out the characteristics and relationships of each image-text pair, we first calculate relative distances between positive and negative pairs in the anchor branch, which serves as a prior and valuable insight into sample relationships of the original single branch. With the learned distance from the anchor branch, we introduce an adaptive margin for each triplet and impose more plausible restrictions when training the target branch, which we define as \textit{Relative Sum (RS)} by:
\begin{equation}
\label{eq:relsum}
\begin{split}
\ell_{RS}^{\prime} = 
&\sum\nolimits_{\grave{\boldsymbol{c}}}
[\gamma + (\mathcal{S}_{\boldsymbol{i}, \boldsymbol{c}}^{a} -
\mathcal{S}_{\boldsymbol{i}, \grave{\boldsymbol{c}}}^{a}) - (\mathcal{S}_{\boldsymbol{i}, \boldsymbol{c}}^{t} - \mathcal{S}_{\boldsymbol{i}, \grave{\boldsymbol{c}}}^{t})]_{+} \\
+ &\sum\nolimits_{\grave{\boldsymbol{i}}}
[\gamma + (\mathcal{S}_{\boldsymbol{i}, \boldsymbol{c}}^{a} -
\mathcal{S}_{\grave{\boldsymbol{i}}, \boldsymbol{c}}^{a}) - (\mathcal{S}_{\boldsymbol{i}, \boldsymbol{c}}^{t} - \mathcal{S}_{\grave{\boldsymbol{i}}, \boldsymbol{c}}^{t})]_{+} .
\end{split}
\end{equation}
With well-founded priori from the anchor branch, the goal of $\ell_{RS}^{\prime}$ is to implement adaptive penalties on the triplets, sufficiently pulling apart the relative distances while ensuring the stability of network convergence. However, a crucial caveat of the boosting strategy is the mining of hard negatives, as otherwise the training process will suffer from moderate negatives and will quickly stagnate. This is inspired by the analysis of minimizing a modified non-trivial loss function with uniform sampling in classification tasks~\cite{Detection:DTPBM,Detection:exemplarSVMs}, identification tasks~\cite{ReID:HardTriplet,ReID:Quadruplet}, and multi-modal tasks~\cite{ITM:MPL,ITM:AOQ}. To emphasize hardest negatives ($\check{\boldsymbol{c}},\check{\boldsymbol{i}}$) for each positive pair ($\boldsymbol{i},\boldsymbol{c}$), we formulate \textit{Relative Max (RM)} as:
\begin{equation}
\label{eq:relmax}
\begin{split}
\ell_{RM}^{\prime} = 
& [\gamma +  (\mathcal{S}_{\boldsymbol{i}, \boldsymbol{c}}^{a} -
\mathcal{S}_{\boldsymbol{i}, \check{\boldsymbol{c}}}^{a})  - (\mathcal{S}_{\boldsymbol{i}, \boldsymbol{c}}^{t} - \mathcal{S}_{\boldsymbol{i}, \check{\boldsymbol{c}}}^{t})]_{+} \\
+ & [\gamma + (\mathcal{S}_{\boldsymbol{i}, \boldsymbol{c}}^{a} -
\mathcal{S}_{\check{\boldsymbol{i}}, \boldsymbol{c}}^{a}) - (\mathcal{S}_{\boldsymbol{i}, \boldsymbol{c}}^{t} - \mathcal{S}_{\check{\boldsymbol{i}}, \boldsymbol{c}}^{t})]_{+} ,
\end{split}
\end{equation}
where
$\check{\boldsymbol{i}}={\argmax}_{\grave{\boldsymbol{i}}}(\mathcal{S}_{\grave{\boldsymbol{i}},\boldsymbol{c}}^{t} - \mathcal{S}_{\grave{\boldsymbol{i}},\boldsymbol{c}}^{a})$ and $\check{\boldsymbol{c}}={\argmax}_{\grave{\boldsymbol{c}}}(\mathcal{S}_{\boldsymbol{i},\grave{\boldsymbol{c}}}^{t} - \mathcal{S}_{\boldsymbol{i},\grave{\boldsymbol{c}}}^{a})$. It is worth noting that ($\check{\boldsymbol{i}},\check{\boldsymbol{c}}$) represent the unmatched samples where the relative distances in the target branch are the toughest to push away based on the ones in the anchor branch, rather than the most confusing negatives ($\hat{\boldsymbol{i}},\hat{\boldsymbol{c}}$) in the target branch itself. At this point, the training difficulty of the target branch is relatively higher aiming to further capture discriminative matching details and improve the quality of the learned metrics. Even with its effectiveness, we argue that the relative boosting strategy does not specify how close the positive pairs are and how far the negative pairs are, inevitably rendering an insufficient exploration of positive-pair intimacy and negative-pair alienation.

\begin{figure*}[t!]
	\centering
	\begin{tabular}{@{}c}
		\includegraphics[width=0.98\linewidth, height=0.28\linewidth,trim= 0 310 185 0,clip]{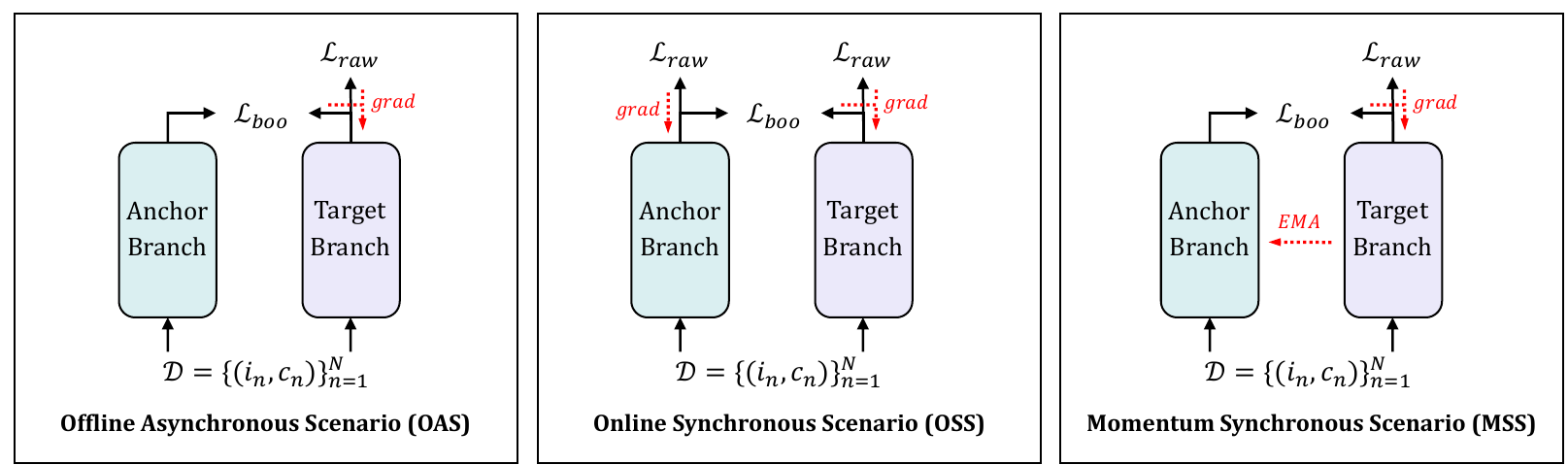} 
	\end{tabular}
	\caption{Illustration of multiple training scenarios. OAS adopts two-stage training scheme as with Conventional Distillation, while OSS and MSS employ one-stage parallel training scenario as with Mutual Learning and Contrastive Learning, respectively. Notably, we only verify the target branch on the validation set, and utilize the model with the best RSUM to perform prediction on the test set.}
	\label{fig:MTS}
\end{figure*}

\textbf{Absolute Boosting Strategy.} Based on the above observation, we explicitly normalize the absolute distances by comparing positive or negative pairs respectively between target and anchor branches. Hence, in contrast to the relative strategy, we directly adjust two new adaptive and explicit margins for the matched and unmatched pairs to pull the former closer, and meanwhile push the latter farther from each other in the target branch. Translating this statement into equation, we define \textit{Absolute Sum (AS)} as:
\begin{equation}
\label{eq:abssum}
\begin{split}
\ell_{AS}^{\prime} =
&\sum\nolimits_{\grave{\boldsymbol{c}}}
([\gamma_{1} \!+\! \mathcal{S}_{\boldsymbol{i}, \boldsymbol{c}}^{a} 
\!-\! \mathcal{S}_{\boldsymbol{i}, \boldsymbol{c}}^{t}]_{+} + 
[\gamma_{2} \!+\! \mathcal{S}_{\boldsymbol{i}, \grave{\boldsymbol{c}}}^{t}
\!-\! \mathcal{S}_{\boldsymbol{i}, \grave{\boldsymbol{c}}}^{a}]_{+}) \\
+ &\sum\nolimits_{\grave{\boldsymbol{i}}}
([\gamma_{1} \!+\! \mathcal{S}_{\boldsymbol{i}, \boldsymbol{c}}^{a}
\!-\! \mathcal{S}_{\boldsymbol{i}, \boldsymbol{c}}^{t}]_{+} +  
[\gamma_{2} \!+\! \mathcal{S}_{\grave{\boldsymbol{i}}, \boldsymbol{c}}^{t}
\!-\! \mathcal{S}_{\grave{\boldsymbol{i}}, \boldsymbol{c}}^{a}]_{+}) ,
\end{split}
\end{equation}
where $\gamma_{1}=\alpha\gamma$, $\gamma_{2}=\gamma-\alpha\gamma$ (consistent with Eq.~\eqref{eq:sumlos}\eqref{eq:maxlos}\eqref{eq:relsum}). As described above, we also exploit hard negative mining to discover the hidden details between image regions and text words, and produce larger gaps between positives and negatives. The \textit{Absolute Max (AM)} can be formulated as:
\begin{equation}
\label{eq:absmax}
\begin{split}
\ell_{AM}^{\prime} =
&[\gamma_{1} + \mathcal{S}_{\boldsymbol{i}, \boldsymbol{c}}^{a} - \mathcal{S}_{\boldsymbol{i}, \boldsymbol{c}}^{t}]_{+} +  
[\gamma_{2} + \mathcal{S}_{\boldsymbol{i}, \check{\boldsymbol{c}}}^{t} - \mathcal{S}_{\boldsymbol{i}, \check{\boldsymbol{c}}}^{a}]_{+} \\
+ &[\gamma_{1} + \mathcal{S}_{\boldsymbol{i}, \boldsymbol{c}}^{a} - \mathcal{S}_{\boldsymbol{i}, \boldsymbol{c}}^{t}]_{+} +  
[\gamma_{2} + \mathcal{S}_{\check{\boldsymbol{i}}, \boldsymbol{c}}^{t} - \mathcal{S}_{\check{\boldsymbol{i}}, \boldsymbol{c}}^{a}]_{+} ,
\end{split}
\end{equation}
where the hardest negatives ($\check{\boldsymbol{i}}, \check{\boldsymbol{c}}$) are mathematically equivalent to the ones of Eq.~\eqref{eq:relmax} in the same mini-batch. Different from the relative loss function that only requires the distance difference between anchor and target branches to be less than a unified margin, the absolute loss function attempts to simultaneously impose explicit and tighter penalties on absolute distances of positive and negative pairs. By this means, the latter can further enhance the discriminative power of the target branch, and develop the optimal feature and distance metric jointly for image-text matching.

\textbf{Discussion.} 
We utilize the relative strategy to only supervise the relative distance within triplets, while the absolute strategy further constrains the absolute distance among each pair.

\noindent\textbf{1) Relative vs. Absolute.} To take $\ell_{RM}^{\prime}$ and $\ell_{AM}^{\prime}$ as an example, we derive their connections by the formulas:
\begin{equation}
\label{eq:i2t}
\begin{split}
[\gamma + &(\mathcal{S}_{\boldsymbol{i}, \boldsymbol{c}}^{a} 
- \mathcal{S}_{\boldsymbol{i}, \check{\boldsymbol{c}}}^{a})  - (\mathcal{S}_{\boldsymbol{i}, \boldsymbol{c}}^{t} - \mathcal{S}_{\boldsymbol{i}, \check{\boldsymbol{c}}}^{t})]_{+} \\
&=[(\gamma_{1} + \mathcal{S}_{\boldsymbol{i}, \boldsymbol{c}}^{a} - \mathcal{S}_{\boldsymbol{i}, \boldsymbol{c}}^{t}) +  
(\gamma_{2} + \mathcal{S}_{\boldsymbol{i}, \check{\boldsymbol{c}}}^{t} - \mathcal{S}_{\boldsymbol{i}, \check{\boldsymbol{c}}}^{a})]_{+} \\
&\leq [\gamma_{1} + \mathcal{S}_{\boldsymbol{i}, \boldsymbol{c}}^{a} - \mathcal{S}_{\boldsymbol{i}, \boldsymbol{c}}^{t}]_{+} +  
[\gamma_{2} + \mathcal{S}_{\boldsymbol{i}, \check{\boldsymbol{c}}}^{t} - \mathcal{S}_{\boldsymbol{i}, \check{\boldsymbol{c}}}^{a}]_{+} ,
\end{split}
\end{equation}
\begin{equation}
\label{eq:t2i}
\begin{split}
[\gamma + &(\mathcal{S}_{\boldsymbol{i}, \boldsymbol{c}}^{a} -\mathcal{S}_{\check{\boldsymbol{i}}, \boldsymbol{c}}^{a}) - (\mathcal{S}_{\boldsymbol{i}, \boldsymbol{c}}^{t} - \mathcal{S}_{\check{\boldsymbol{i}}, \boldsymbol{c}}^{t})]_{+} \\
&=(\gamma_{1} + \mathcal{S}_{\boldsymbol{i}, \boldsymbol{c}}^{a} - \mathcal{S}_{\boldsymbol{i}, \boldsymbol{c}}^{t}) +  
(\gamma_{2} + \mathcal{S}_{\check{\boldsymbol{i}}, \boldsymbol{c}}^{t} - \mathcal{S}_{\check{\boldsymbol{i}}, \boldsymbol{c}}^{a})]_{+} \\
&\leq[\gamma_{1} + \mathcal{S}_{\boldsymbol{i}, \boldsymbol{c}}^{a} - \mathcal{S}_{\boldsymbol{i}, \boldsymbol{c}}^{t}]_{+} +  
[\gamma_{2} + \mathcal{S}_{\check{\boldsymbol{i}}, \boldsymbol{c}}^{t} - \mathcal{S}_{\check{\boldsymbol{i}}, \boldsymbol{c}}^{a}]_{+} . \\
    \end{split}
\end{equation}

Combining Eq.~\eqref{eq:i2t} and~\eqref{eq:t2i}, we can obtain the inequality relations between two boosting strategies as follows:
\begin{equation}
    \ell_{RM}^{\prime} \leq \ell_{AM}^{\prime} ,
\end{equation}
where the equality holds if and only if three items
$(\gamma_{1} + \mathcal{S}_{\boldsymbol{i}, \boldsymbol{c}}^{a} - \mathcal{S}_{\boldsymbol{i}, \boldsymbol{c}}^{t}),
(\gamma_{2} + \mathcal{S}_{\boldsymbol{i}, \check{\boldsymbol{c}}}^{t} - \mathcal{S}_{\boldsymbol{i}, \check{\boldsymbol{c}}}^{a}),
(\gamma_{2} + \mathcal{S}_{\check{\boldsymbol{i}}, \boldsymbol{c}}^{t} - \mathcal{S}_{\check{\boldsymbol{i}}, \boldsymbol{c}}^{a})$ share the same signs. Similarly, $\ell_{RS}^{\prime} \leq \ell_{AS}^{\prime}$, confirming that the absolute form can impose tighter constraints, and produce more compact distances than the relative one (See Sec.~\ref{subsecABS}).

\noindent\textbf{2) Fixed $\gamma$ vs. Soft ${\gamma}^{SA}$.}
The standard configurations in Eq.~\eqref{eq:relsum}\eqref{eq:relmax}\eqref{eq:abssum}\eqref{eq:absmax} recommend RM and AM with fixed $\gamma$, meaning that the distance metric learned by the target branch has a consistent $\gamma$ penalty based on the ones learned by the anchor branch. On the other hand, there is a value range of all available measures used for boosting strategies, e.g. the absolute distance of postive/negative pairs  $\mathcal{S}_{\boldsymbol{i},\boldsymbol{c}}^{a}/\mathcal{S}_{\boldsymbol{i},\check{\boldsymbol{c}}}^{a}/\mathcal{S}_{\check{\boldsymbol{i}},\boldsymbol{c}}^{a}\in[-d_{y},d_{y}]_{d_{y}=1}$ in Eq.~\eqref{eq:absmax}, and the relative distance within each triplet $(\mathcal{S}_{\boldsymbol{i},\boldsymbol{c}}^{a}-\mathcal{S}_{\boldsymbol{i},\check{\boldsymbol{c}}}^{a})/(\mathcal{S}_{\boldsymbol{i},\boldsymbol{c}}^{a}-\mathcal{S}_{\check{\boldsymbol{i}},\boldsymbol{c}}^{a})\in[-d_{x},d_{x}]_{d_{x}=2}$ in Eq.~\eqref{eq:relmax}. In other words, each type of the above distances theoretically has a boosting extreme namely \textbf{Theoretical Maximum (TM)}, some of which are less than the predefined $\gamma$. Combining these two findings, we propose a soft $\gamma^{SA}$ formulation namely \textbf{Soft Adaptation (SA)} which varies exponentially corresponding to predicted distance from the anchor branch.

For $\gamma$ in RM, we initialize $\gamma^{SA}(x)$ that corresponds to the output score from the anchor branch as follows:
\begin{equation}
\gamma^{SA}(x) = \frac{2\gamma}{1+e^{\epsilon(x-d_{x})}}-\gamma ,
\end{equation}
where $\epsilon$ is a smooth value that controls the sharpness of the curve $\gamma^{SA}(x)$. $x$ indicates the inferred relative distance within each triplet from anchor branch. To maximize margin penalty without exceeding the extreme, we obtain an equation as:
\begin{equation}
\left.{\frac{\partial\gamma^{SA}(x)}{\partial x}}\right|_{x=d_{x}} = \left.{\frac{\partial\gamma^{TM}(x)}{\partial x}}\right|_{x=d_{x}} .
\end{equation}
Given the above equation, we then compute the $\epsilon$ value. The derivation is as follows:
\begin{equation}
    \begin{split}
    \left.{\frac{\partial\gamma^{SA}(x)}{\partial x}}\right|_{x=d_{x}} &=
    \left.\frac{-2\gamma\epsilon e^{\epsilon(x-d_{x})}}{(1+e^{\epsilon(x-d_{x})})^2}\right|_{x=d_{x}} = \frac{-\gamma\epsilon}{2} , \\
    \left.{\frac{\partial\gamma^{TM}(x)}{\partial x}}\right|_{x=d_{x}} &= \left.{\frac{\partial{(d_{x}-x)}}{\partial x}}\right|_{x=d_{x}} = -1 .
    \end{split}
\end{equation}
\noindent Hence, $\epsilon=\frac{2}{\gamma}$. The $\gamma^{SA}(x)$ for RM can be formulated as:
\begin{equation}
\gamma^{SA}(x) = \frac{2\gamma}{1+e^{\frac{2}{\gamma}(x-d_{x})}}-\gamma
=\frac{\gamma-\gamma e^{\frac{2}{\gamma}(x-d_{x})}}{1+e^{\frac{2}{\gamma}(x-d_{x})}} \ ,
\end{equation}
and we redefine the \textit{Relative Max (RM)} as:
\begin{equation}
\label{eq:softrelmax}
\begin{split}
\ell_{RM}^{\prime} =\
&[\gamma^{SA}(\mathcal{S}_{\boldsymbol{i}, \boldsymbol{c}}^{a} \!-\!
\mathcal{S}_{\boldsymbol{i}, \check{\boldsymbol{c}}}^{a}) \!+\!  (\mathcal{S}_{\boldsymbol{i}, \boldsymbol{c}}^{a} \!-\!
\mathcal{S}_{\boldsymbol{i}, \check{\boldsymbol{c}}}^{a})  \!-\! (\mathcal{S}_{\boldsymbol{i}, \boldsymbol{c}}^{t} \!-\! 
\mathcal{S}_{\boldsymbol{i}, \check{\boldsymbol{c}}}^{t})]_{+} \\
+\ &[\gamma^{SA}(\mathcal{S}_{\boldsymbol{i}, \boldsymbol{c}}^{a} \!-\! \mathcal{S}_{\check{\boldsymbol{i}}, \boldsymbol{c}}^{a}) \!+\! (\mathcal{S}_{\boldsymbol{i}, \boldsymbol{c}}^{a} \!-\!
\mathcal{S}_{\check{\boldsymbol{i}}, \boldsymbol{c}}^{a}) \!-\! (\mathcal{S}_{\boldsymbol{i}, \boldsymbol{c}}^{t} \!-\! \mathcal{S}_{\check{\boldsymbol{i}}, \boldsymbol{c}}^{t})]_{+} .
\end{split}
\end{equation}

Similarly for $\gamma_{1},\gamma_{2}$ in AM, we obtain $\gamma_{1}^{SA}(y),\gamma_{2}^{SA}(y)$ as:
\begin{equation}
\begin{split}
\gamma_{1}^{SA}(y) &= \frac{2\gamma_{1}}{1+e^{\frac{2}{\gamma_{1}}(y-d_{y})}}-\gamma_{1}
=\frac{\gamma_{1}-\gamma_{1} e^{\frac{2}{\gamma_{1}}(y-d_{y})}}{1+e^{\frac{2}{\gamma_{1}}(y-d_{y})}} ,\\
\gamma_{2}^{SA}(y) &= \frac{2\gamma_{2}}{1+e^{\frac{-2}{\gamma_{2}}(y+d_{y})}}-\gamma_{2}
=\frac{\gamma_{2}-\gamma_{2} e^{\frac{-2}{\gamma_{2}}(y+d_{y})}}{1+e^{\frac{-2}{\gamma_{2}}(y+d_{y})}} ,
\end{split}
\end{equation}
and we reformulate the \textit{Absolute Max (AM)} as:
\begin{equation}
\label{eq:softabsmax}
\begin{split}
\ell_{AM}^{\prime} =\
&[\gamma_{1}^{SA}(\mathcal{S}_{\boldsymbol{i}, \boldsymbol{c}}^{a}) \!+\! \mathcal{S}_{\boldsymbol{i}, \boldsymbol{c}}^{a} \!-\!
\mathcal{S}_{\boldsymbol{i}, \boldsymbol{c}}^{t}]_{+}
\!+\! [\gamma_{2}^{SA}(\mathcal{S}_{\boldsymbol{i}, \check{\boldsymbol{c}}}^{a}) \!+\! \mathcal{S}_{\boldsymbol{i}, \check{\boldsymbol{c}}}^{t} \!-\!
\mathcal{S}_{\boldsymbol{i}, \check{\boldsymbol{c}}}^{a}]_{+} \\
+\ &[\gamma_{1}^{SA}(\mathcal{S}_{\boldsymbol{i}, \boldsymbol{c}}^{a}) \!+\! \mathcal{S}_{\boldsymbol{i}, \boldsymbol{c}}^{a} \!-\!
\mathcal{S}_{\boldsymbol{i}, \boldsymbol{c}}^{t}]_{+}
\!+\! [\gamma_{2}^{SA}(\mathcal{S}_{\check{\boldsymbol{i}}, \boldsymbol{c}}^{a}) \!+\! \mathcal{S}_{\check{\boldsymbol{i}}, \boldsymbol{c}}^{t} \!-\! \mathcal{S}_{\check{\boldsymbol{i}}, \boldsymbol{c}}^{a}]_{+} .
\end{split}
\end{equation}
Experiments in Sec.~\ref{subsecABS} show that the soft one displays stronger abilities of image retrieval but slightly attractive promotions on sentence retrieval, while the fixed one obtains the optimal balance between bidirectional retrievals. In summary, we recommend the fixed one as the vanilla boosting strategy.

\subsection{Multiple Training Scenarios}
\label{subsec:MTS}
The collaboration of target and anchor branches is flexible. Following Conventional Distillation~\cite{KD:KD,KD:SP,KD:RKD,KD:PKT}, Mutual Learning~\cite{KD:DML}, and Contrastive Learning~\cite{CL:MoCo,CL:DINO,CL:BYOL}, we adopt three popular and corresponding scenarios namely Offline Asynchronous, Online Synchronous and Momentum Synchronous Scenarios to validate the effectiveness and generalization of our boosting strategy.

\textbf{Conventional Distillation.} 
Traditional knowledge distillation~\cite{KD:KD,KD:Darkrank,KD:SP,KD:RKD} is a mechanism where the target (student) branch learns to match the results of the static anchor (teacher) branch, parameterized by $\theta_{t}$ and $\theta_{a}$ respectively:
\begin{equation}
    \min\nolimits_{\theta_{t}} \mathcal{L}_{kd}(F_{\theta_{t}}(x),F_{\theta_{a}}(x)) \ ,
\end{equation}
where $\mathcal{L}_{kd}$ consists of logit/distance/angle-wise distillation loss functions which help the target branch transfer the powerful knowledge from the pre-trained anchor branch. To evaluate it, we first obtain a strong anchor branch by task-specific loss function $\mathcal{L}_{raw}$. As a post-processing step, the training procedure of the target branch is then supervised by $\mathcal{L}_{raw}+\mathcal{L}_{boo}$ which penalizes the proximities between two branches and agitates the latter to gain the better matching ability, denoted as \textit{Offline Asynchronous Scenario (OAS)}.

\textbf{Mutual Learning.}
Unlike general knowledge distillation, the branch cohorts are updated jointly and collaboratively in deep mutual learning which does not rely on prior knowledge. For example, DML~\cite{KD:DML} attempts to optimize the cohorts by bringing their probability estimates closer and minimizing their discrepancies during the learning progress. However, our proposed approach aims to implicitly push aside the peers’ distributions and exploit more powerful paradigms beyond mimicry. Hence, simultaneously using boosting strategy for each branch may lead to unclear training objectives and unstable optimizing paths, and ultimately converge to locally mediocre solutions for peer training. Therefore, we randomly initialize two branches where only the target branch is updated under the guidance of the boosting strategy by the anchor branch, denoted as \textit{Online Synchronous Scenario (OSS)}.

\textbf{Contrastive Learning.}
To avoid model collapse, several works focus on contrastive loss~\cite{CL:NCELoss}, inconsistent structure~\cite{CL:BYOL,CL:SimSiam}, clustering constraint~\cite{CL:DeepCluster,CL:swAV}, and momentum encoder~\cite{CL:MoCo,CL:DINO}. Likewise, they do not require a pre-trained network given a priori. Inspired by the momentum encoder~\cite{CL:MoCo}, we update the anchor parameters $\theta_{a}$ with an exponential moving average (EMA) of the target parameters $\theta_{t}$. Formally, we update $\theta_{a}$ by:
\begin{equation}
    \theta_{a} \gets \beta\theta_{a}+(1-\beta)\theta_{t} ,
\end{equation}
where $\beta$ follows a cosine schedule~\cite{CL:DINO,CL:BYOL} from 0.99995 to 1 during training process. Only the target branch is updated by back-propagation of $\mathcal{L}_{raw}+\mathcal{L}_{boo}$, and the dynamic anchor branch progressively provides a consistent reference of higher quality and hence, we have no need of maintaining the queue dictionary of data samples or introducing data augmentation to form positive pairs. Note that this cooperative strategy serves as a standard operation similar to Polyak-Ruppert averaging with exponential decay~\cite{CL:Polyak,CL:Robbins-Monro}, which is denoted as \textit{Momentum Synchronous Scenario (MSS)}.

\textbf{Discussion.} Deep cooperative learning is a popular technique trading extra training consumption for performance gains, and only utilizing the target branch for prediction under the above training scenarios. OAS adopts a two-stage training scheme as with DR~\cite{KD:Darkrank} and RKD~\cite{KD:RKD}, while OSS and MSS train two branches simultaneously at one stage as with DML~\cite{KD:DML} and DINO~\cite{CL:DINO} respectively. Compared with OSS, OAS and MSS require no gradients for the anchor branch during the cooperative process. We directly employ $\mathcal{L}_{raw}$ and $\mathcal{L}_{boo}$ with a 1:1 contribution to train the target branch, which has achieved steady and consistent improvements in all experiments without complex manual tuning. For a fair comparison, we ensure the same training and inference expenses, and validate that MSS can obtain great benefits with both slight training time and memory costs in TABLE~\ref{tab:CooLea}.

\section{Experiments}
We first introduce the detailed training settings. Then, we report the cooperation and comparison with recent works and some popular strategies. After that, we investigate the configurations and analyses of our proposed DBL. Finally, we visualize some illustrations of bidirectional retrieval examples.

\subsection{Datasets and Settings}

\textbf{Benchmark Datasets.}
We evaluate our proposed approaches on two benchmark datasets: Flickr30K~\cite{Datasets:Flickr30k} and MSCOCO~\cite{Datasets:MSCOCO}. Each image of these two datasets is annotated with five corresponding captions. For Flickr30K, we adopt the standard split~\cite{ITM:DeViSE} and divide the dataset into 29,000 training images, 1,000 validation images, and 1,000 testing images. For MSCOCO, we follow~\cite{ITM:VSE++,ITM:SCAN} to utilize 113,287 images for training, 5,000 images for validation and 5,000 images for testing. The 1k evaluation result is computed by averaging over 5 folds of 1K test images on MSCOCO.

\textbf{Evaluation Metrics.}
We adopt Recall@$\kappa$ (R@$\kappa$) and sum (RSUM) of R@1, R@5, and R@10 in two directions for evaluation, where R@$\kappa$ indicates the percentage of queries whose correct response is included in the top-$\kappa$ candidates. Since R@$\kappa$ evaluation only cares about the first groundtruth retrieved in the top-$\kappa$ results, we introduce the mean distance (MD) between positive and negative candidates for a better illustration of the similarity separability.

\textbf{Implementation Details.}
Inspired by similarity representations~\cite{ITM:GSMN,ITM:SGRAF,ITM:CMCAN}, our baseline is an improved version of SCAN~\cite{ITM:SCAN} with the vectorized similarity~\cite{ITM:SGRAF} instead of cosine distance. Following them, we extract $K$=36 salient regions by bottom-up attention~\cite{IC:BU_TDA} for each image, and map 300-d word embeddings with random initialization into 1024-d features by Bi-GRU. We set the dimension of alignment vectors to be 256 with the inversed temperature $\lambda$ as 9. The margin value $\gamma$ and the proportion $\alpha$ are set as 0.2 and 0.5 respectively. We train our method with 20 and 40 epochs on MSCOCO and Flickr30k dataset respectively during all the training scenarios, and set the initial learning rate as 0.0002 for 10 and 30 epochs, and decay it by 0.1 for the rest epochs.

\begin{table*}[t!]
    \caption{Cooperation with multiple representative models including Embedding-based and Interaction-based learning on Flickr30K and MSCOCO. We re-implement these methods with their publicly available code. The best results of the RSUM are marked in \textbf{bold}.}\label{tab:cocof30k}
    \centering
    \setlength{\tabcolsep}{0.9mm}{
    \begin{tabular}{clccccccccccccccc}
        \hline
        &\multirow{3}{*}{\bf Method}
        &\multicolumn{4}{c} {Flickr30K} & 
        &\multicolumn{4}{c} {MSCOCO 1K} &
        &\multicolumn{4}{c} {MSCOCO 5K} &\\
        \cline{3-6} \cline{8-11} \cline{13-16}
        &&\multicolumn{2}{c}{Sentence Retrieval} 
        &\multicolumn{2}{c}{Image Retrieval} &
        &\multicolumn{2}{c}{Sentence Retrieval} 
        &\multicolumn{2}{c}{Image Retrieval} &
        &\multicolumn{2}{c}{Sentence Retrieval} 
        &\multicolumn{2}{c}{Image Retrieval} &	\\
        &&R@1 &R@5 &R@1 &R@5 &RSUM
        &R@1 &R@5 &R@1 &R@5 &RSUM
        &R@1 &R@5 &R@1 &R@5 &RSUM \\
        \hline
        \multirow{9}{*}{\rotatebox{90}{\textbf{Embedding Learning}}}
        &\textbf{VSRN~\cite{ITM:VSRN}}
        &70.2 &89.4 &53.2 &78.0 &471.2
        &74.2 &94.1 &60.6 &88.3 &509.2
        &50.3 &79.4 &37.6 &68.5 &403.3
        \\
        &+RM (OSS)
        &72.1 &90.3 &54.8 &78.6 &476.3
        &75.3 &94.6 &61.5 &89.0 &512.8
        &51.6 &79.9 &38.8 &69.5 &407.4\\
        &+AM (OSS)
        &72.8 &90.4 &55.0 &78.9 &\textbf{477.1}
        &75.2 &94.8 &61.8 &89.2 &\textbf{513.2}
        &51.7 &80.0 &39.4 &69.7 &\textbf{408.6}\\
        \cline{2-17}
        &\textbf{ESA~\cite{ITM:ESA}}
        &82.3 &95.8 &61.2 &86.0 &514.5
        &79.2 &96.4 &63.5 &90.8 &524.9
        &58.0 &84.8 &41.2 &71.3 &429.3\\
        &+RM (OSS)
        &83.4 &96.1 &61.9 &86.4 &517.3
        &80.0 &96.5 &63.7 &91.1 &526.3
        &58.6 &85.0 &41.5 &72.0 &431.2\\
        &+AM (OSS)
        &83.2 &96.2 &62.2 &86.5 &\textbf{517.5}
        &80.1 &96.5 &63.8 &91.2 &\textbf{526.7}
        &58.8 &85.2 &41.6 &72.0 &\textbf{431.8}\\
        \cline{2-17}
        &\textbf{CLIP~\cite{VLP:CLIP}}
        &92.0 &99.4 &77.8 &95.0 &561.1
        &83.2 &96.4 &68.3 &91.3 &534.8
        &67.5 &88.3 &49.4 &75.1 &457.8\\
        &+RM (OAS)
        &93.0 &99.5 &79.1 &95.3 &\textbf{563.8}
        &83.9 &96.8 &68.7 &91.6 &536.5
        &68.2 &88.6 &49.8 &75.3 &459.7\\
        &+AM (OAS)
        &92.8 &99.5 &78.9 &95.2 &563.3
        &84.1 &97.0 &68.8 &91.6 &\textbf{536.7}
        &68.3 &88.8 &50.1 &75.4 &\textbf{460.1}\\
        \hline
        \multirow{12}{*}{\rotatebox{90}{\textbf{Interaction Learning}}}
        &\textbf{BFAN~\cite{ITM:BFAN}}
        &70.3 &91.6 &53.2 &78.5 &474.2
        &75.5 &94.3 &60.8 &88.0 &510.6
        &54.6 &82.1 &38.8 &68.7 &413.8\\
        &+RM (OSS)
        &72.8 &93.3 &55.5 &79.3 &\textbf{483.5}
        &77.9 &95.8 &62.4 &89.2 &\textbf{518.7}
        &56.4 &83.6 &40.7 &69.8 &\textbf{421.9}\\
        &+AM (OSS)
        &73.9 &92.6 &55.3 &78.6 &481.0
        &77.7 &95.7 &62.4 &89.1 &518.0
        &55.7 &83.6 &40.4 &70.0 &420.9\\
        \cline{2-17}
        &\textbf{SGRAF~\cite{ITM:SGRAF}}
        &78.1 &94.4 &58.2 &83.1 &500.2
        &79.3 &96.2 &63.5 &90.3 &523.8
        &58.1 &85.0 &41.8 &71.3 &429.6\\
        &+RM (OSS)
        &80.1 &94.7 &59.6 &83.5 &504.9
        &79.8 &96.7 &64.0 &90.5 &525.5
        &59.2 &84.9 &42.3 &71.6 &432.2\\
        &+AM (OSS)
        &79.8 &95.4 &60.7 &84.0 &\textbf{507.2}
        &80.5 &96.6 &64.3 &90.6 &\textbf{526.5}
        &59.8 &85.2 &42.5 &71.5 &\textbf{433.1}\\
        \cline{2-17}
        &\textbf{NAAF~\cite{ITM:NAAF}}
        &78.3 &96.1 &59.6 &84.4 &506.6
        &77.8 &95.8 &62.5 &89.6 &519.5
        &56.3 &84.1 &40.9 &70.1 &422.9\\
        &+RM (OSS)
        &78.6 &96.2 &60.1 &84.4 &507.8
        &78.4 &96.1 &63.1 &89.6 &521.0
        &56.8 &84.2 &41.3 &70.2 &\textbf{424.3}\\
        &+AM (OSS)
        &79.1 &96.4 &60.3 &84.6 &\textbf{509.1}
        &78.8 &95.9 &62.9 &89.5 &\textbf{521.4}
        &57.2 &84.0 &41.1 &69.8 &424.0\\
        \cline{2-17}
        &\textbf{OSCAR~\cite{VLP:Oscar}}
        &-- &-- &-- &-- &-- 
        &88.4 &99.1 &75.7 &95.2 &556.5 
        &70.0 &91.1 &54.0 &80.8 &479.9\\
        &+RM (OAS)
        &-- &-- &-- &-- &--
        &88.8 &98.9 &75.8 &95.4 &557.1
        &71.0 &91.0 &54.5 &80.9 &481.2\\
        &+AM (OAS)
        &-- &-- &-- &-- &-- 
        &88.8 &99.0 &76.1 &95.5 &\textbf{557.4}
        &70.9 &91.1 &54.8 &81.0 &\textbf{481.4}\\
        \hline
    \end{tabular}}
\end{table*}

\subsection{Quantitative Results}
\label{subsec:QR}

\textbf{Cooperation with Multiple Models.}
TABLE~\ref{tab:cocof30k} lists the applications on two types of image-text matching architectures, including Embedding-based (VSRN~\cite{ITM:VSRN}, ESA~\cite{ITM:ESA}, CLIP~\cite{VLP:CLIP}) and Interaction-based (BFAN~\cite{ITM:BFAN}, SGRAF~\cite{ITM:SGRAF}, NAAF~\cite{ITM:NAAF}, OSCAR~\cite{VLP:Oscar}) methods. 
Considering the diverse settings of these methods, it is essential to mention that we uniformly adopt the setups of BiGRU and a single model based on VSRN, ESA, BFAN, and NAAF. 
In particular, we utilize BFAN with equal attention and similarity representation, CLIP with ViT-L/14@336px encoder, and OSCAR with base BERT as references.
Limitted by current resources, we apply DBL under OSS for relatively smaller VSRN, ESA, BFAN, SGRAF, and NAAF, and under OAS for larger pre-trained CLIP and OSCAR.
We keep their original loss functions, training settings, and model configurations. It is worth noting that we only utilize the target branch for performance validation.

\textit{Visual Semantic Reasoning (VSRN)}~\cite{ITM:VSRN} performs graph reasoning to generate visual features with semantic relationships.

\textit{External Space Attention Aggregation (ESA)}~\cite{ITM:ESA} enables element-wise attention for discriminative information and adaptive feature aggregation at the dimensional level.

\textit{Transferable Visual Models From Natural Language Supervision (CLIP)}~\cite{VLP:CLIP} learns dual powerful image and text encoders on massive image-text pairs collected from the internet.

\textit{Bidirectional Focal Attention (BFAN)}~\cite{ITM:BFAN} enhances the region-word correspondence via a focal attention unit, and integrates all region-based and word-based similarities.

\textit{Similarity Graph Reasoning and Attention Filtration (SGRAF)}~\cite{ITM:SGRAF} designs word-based similarity function, followed by graph reasoning and attention filtration modules.

\textit{Negative-aware Attention Framework (NAAF)}~\cite{ITM:NAAF} learns the optimal boundary to explicitly model matched and mismatched fragments, which yield the final score together.

\textit{Object-Semantics Aligned Pre-training (OSCAR)}~\cite{VLP:Oscar} utilizes object tags detected in images as a bridge to considerably alleviate cross-modality gap and alignment burden.

\noindent As can be clearly seen, they all benefit from the DBL strategy at all evaluation metrics for both relative and absolute manners. 
For R@1 at sentence/image retrieval on Flickr30K, it gains a maximal boost of 2.6/1.8\% (VSRN), 1.1/1.0\% (ESA), 3.6/2.3\% (BFAN), 2.0/2.5\% (SGRAF), and 0.8/0.7\% (NAAF) respectively. The impressive and consistent improvements are also shown on MSCOCO 1K and 5K test sets, which well display strong capability and broad applicability regardless of network frameworks.
Besides for pre-trained CLIP, our DBL improves R@1 by at most 1.0/1.3\% and 0.8/0.7\% on Flickr30K and MSCOCO5K, while based on OSCAR, our DBL still obtains a maximum R@1 boost of 1.0/0.8\% on challenging MSCOCO5K.
Notably, we have also attempted to apply DBL to GPO~\cite{ITM:GPO} and found that its warm-up process causes drastic changes in the loss magnitudes (1300-400, 48-14), making it difficult to balance the weights between GPO and DBL. Without a warm-up strategy, GPO with RM and AM can improve the bidirectional R@1, RSUM from 75.3/56.0\%, 493.2\% to 76.8/57.3\%, 499.6\% and 77.2/56.7\%, 500.3\% on Flickr30K, verifying consistent boosts and robustness of our DBL strategy.
In conclusion, it is inadequate for these works supervised by~\cite{ITM:VSE++} to distinguish the relations within triplets, and our DBL can assign adaptive and targeted margins and produce optimal matching patterns between image and text. 

\begin{table}[t!]
    \centering
    \caption{Comparison with cooperative learning on Flickr30K. * indicates the inverse KL divergence of distillation loss.}\label{tab:CooLea}
    \setlength{\tabcolsep}{2pt}{
    \begin{tabular}{lccccccc}
    \hline
    \multirow{2}{*}{Method}
    &\multicolumn{2}{c}{Sentence Retrieval}
    &\multicolumn{2}{c}{Image Retrieval}&
    &\multicolumn{2}{c}{Scenarios}\\
    &R@1&R@5&R@1&R@5&MD&Memory&Time\\
    \hline
    Baseline
    &75.8&93.4&56.5&81.4&0.91&100\%&100\%\\
    \hline
    DR\cite{KD:Darkrank}
    &76.8&93.7&57.4&81.8&1.25
    &\multicolumn{2}{c}{OAS}\\
    RKD\cite{KD:RKD}
    &77.6&93.8&56.8&81.9&0.74
    &\multirow{2}{*}{+12\%} &\multirow{2}{*}{+120\%}\\
    \textbf{RM (OAS)}
    &79.1&\textbf{94.6}&58.5&\textbf{83.3}&1.37\\
    \textbf{AM (OAS)}
    &\textbf{79.7}&\textbf{94.6}&\textbf{58.7}&83.2&\textbf{1.58}\\
    \hline
    DML\cite{KD:DML}
    &77.7&93.6&57.4&82.8&0.94&\multicolumn{2}{c}{OSS}\\
    \textbf{RM (OSS)}
    &\textbf{79.3}&\textbf{95.4}&\textbf{59.1}&83.0&1.21
    &\multirow{2}{*}{+100\%} &\multirow{2}{*}{+73\%}\\
    \textbf{AM (OSS)}
    &79.0&94.9&58.5&\textbf{83.1}&\textbf{1.56}\\
    \hline
    DINO\cite{CL:DINO}
    &76.5&93.4&58.5&83.2&0.89
    &\multicolumn{2}{c}{MSS}\\
    DINO*\cite{CL:DINO}
    &77.9&93.9&58.8&83.5&1.15
    &\multirow{2}{*}{+11\%} &\multirow{2}{*}{+18\%}\\
    \textbf{RM (MSS)}
    &78.4&94.4&59.1&\textbf{83.6}&1.45\\
    \textbf{AM (MSS)}
    &\textbf{79.4}&\textbf{94.7}&\textbf{59.7}&83.5&\textbf{1.81}\\
    \hline
    \end{tabular}}
\end{table}

\textbf{Comparison with Deep Cooperative Learning.}
For fairness, we compare DBL with some typical strategies under their original training settings, including DR (softrank), RKD (distance), DML, and DINO (ema). We extend them with some refinements to be more suitable for cross-modal tasks. For the first two works, we directly utilize the predicted similarity scores for knowledge distillation, while for the last two methods, there are no category labels to supply the imitation process of probability outputs. Hence, we introduce the cross-modal projection~\cite{ITM:CMPL} that treats each image-text pair within a mini-batch as a class and translates the cross-modal scalar projections into the normalized probability estimates between image and text.
Besides, we employ $H(P_{s}(x),P_{t}(x))$ to replace $H(P_{t}(x),P_{s}(x))$ in Eq. (3) of DINO as DINO* according to the observations~\cite{KD:DL,ITM:CMPL} that the latter could blur the distributions of multiple modes and bring ambiguities for image-text matching pattern.
\begin{table}
    \centering
    \caption{Comparison with metric learning on Flickr30K. Here we report MSS due to its close consumption with single branch.}\label{tab:MetLea}
    \setlength{\tabcolsep}{9pt}{
    \begin{tabular}{lccccc}
        \hline
        \multirow{2}{*}{Loss}
        &\multicolumn{2}{c}{Sentence Retrieval}
        &\multicolumn{2}{c}{Image Retrieval}&\\
        &R@1&R@5&R@1&R@5&MD\\
        \hline
        VSE0\cite{ITM:DSPE}
        &74.4&92.8&54.3&81.2&0.58\\
        VSE++\cite{ITM:VSE++}
        &75.8&93.4&56.5&81.4&0.91\\
        SAM\cite{ITM:SAM}
        &76.1&93.2&57.0&82.9&0.88\\
        CMPM\cite{ITM:CMPL}
        &76.8&95.1&57.8&82.8&1.12\\
        MPL\cite{ITM:MPL}
        &77.1&94.3&57.4&82.3&0.96\\
        \hline
        \textbf{RM (MSS)}
        &78.4&94.4&59.1&\textbf{83.6}&1.45\\
        \textbf{AM (MSS)}        
        &\textbf{79.4}&\textbf{94.7}&\textbf{59.7}&83.5&\textbf{1.81}\\
        \hline
    \end{tabular}}
\end{table}
\begin{table}[t!]
    \centering
    \caption{Configurations of boosting loss by OSS on Flickr30K. $\star$ denotes the DBL strategy with soft $\gamma^{SA}$. Both model \#H and model \#I average two boosting loss during training process.}\label{tab:Config}
    \setlength{\tabcolsep}{3pt}{
    \begin{tabular}{lccccccccc}
        \hline
        \multirow{2}{*}{Model}
        &\multicolumn{4}{c}{Strategy}
        &\multicolumn{2}{c}{Sentence Retrieval}
        &\multicolumn{2}{c}{Image Retrieval}&\\
        &RS&RM&AS&AM&R@1&R@5&R@1&R@5&MD\\
        \hline
        A &&&&
        &75.8&93.4&56.5&81.4&0.91\\
        B &\cmark&&&
        &77.6&94.6&58.1&82.4&1.04\\
        C &&&\cmark&
        &78.2&94.8&57.1&82.5&1.13\\
        D &&\cmark$\star$&&
        &77.1&94.5&\textbf{59.6}&\textbf{83.9}&1.23\\
        E &&&&\cmark$\star$
        &77.2&94.6&58.8&83.5&1.30\\
        F &&\cmark&&
        &\textbf{79.3}&\textbf{95.4}&59.1&83.0&1.21\\
        G &&&&\cmark
        &79.0&94.9&58.5&83.1&\textbf{1.56}\\
        H &\cmark&&\cmark&
        &78.0 &94.6 &57.8 &82.3 &1.13\\
        I &&\cmark&&\cmark
        &78.9 &95.1 &58.7 &83.0 &1.44\\
        \hline
        \end{tabular}}
\end{table}
\textbf{1) Performance.} TABLE~\ref{tab:CooLea} shows that our DBL can consistently outweigh them on R@1 by a large margin (Maximum 2.9\% and 1.7\% on sentence and image retrieval), exhibiting good flexibility and broad applicability. Beyond the above, it also indicates that the knowledge transfers in a boosting manner hold tremendous potential for passing messages across branches and significantly promote the matching ability of single-branch learning.
\textbf{2) Efficiency.} OAS/MSS requires no gradients for the anchor branch, and obtains about 12/11\% memory and 120/18\% time increase. Only OSS trains two branches simultaneously with extra 100\% memory and 73\% time costs. Notably, MSS can achieve both slight training time and memory costs.
\textbf{3) Expansion.} Two-branch structure in OAS and MSS is a standard architecture of conventional distillation and contrastive learning. In OSS, DML~\cite{KD:DML} demonstrates another paradigm of multi-branch collaboration for mutual learning. Following it, we extend our strategies to larger peer cohorts.
Concretely, we randomly initialize all branches $\{{B}_{i} \,|\, i=1,...,M\}$ where each branch ${B}_{m}$ is supervised by regarding all previous cohorts $\{{B}_{i} \,|\, i=1,...,m-1\}$ as the anchor branches. 
To ensure comparability between task-specific and boosting losses for current target branch, we average all boosting losses from its corresponding anchor branches, which are then added to the task-specific loss as the final objective function. We discover that the evaluation results slowly converge with the increasing number of peer branches and 3-branch DBL with RM earns the best bidirectional R@1 by 79.4/59.5\% vs. 78.3/58.8\% of 4-branch DML (best). The combination of multiple branches is flexible and two branches can balance complexity and performance.

\begin{figure}[t!]
    \centering
    \begin{tabular}{@{}c}
        \includegraphics[width=0.98 \linewidth, height=0.514 \linewidth,trim= 0 30 0 0,clip]{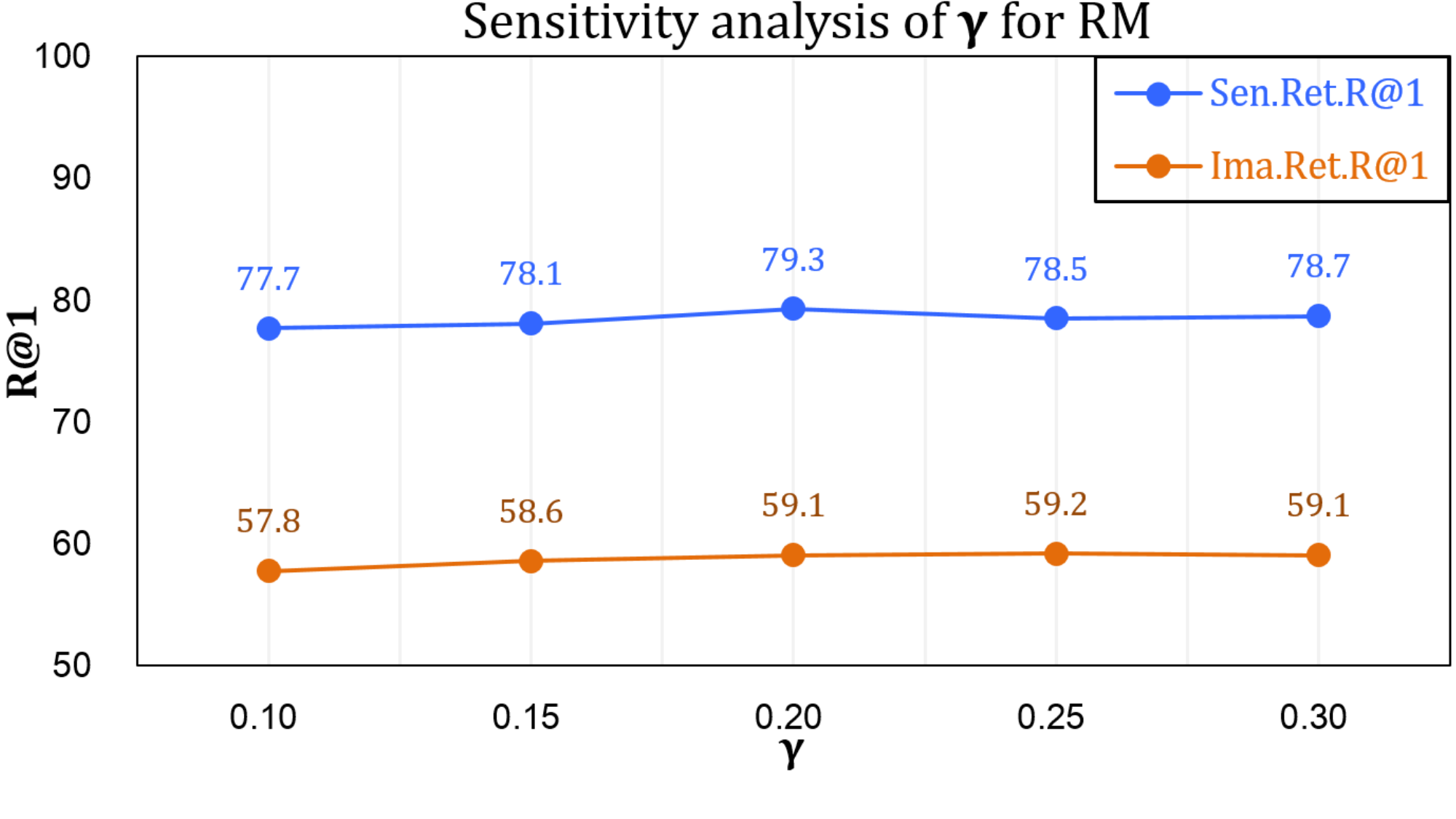} \\
        \includegraphics[width=0.98 \linewidth, height=0.514 \linewidth,trim= 0 30 0 0,clip]{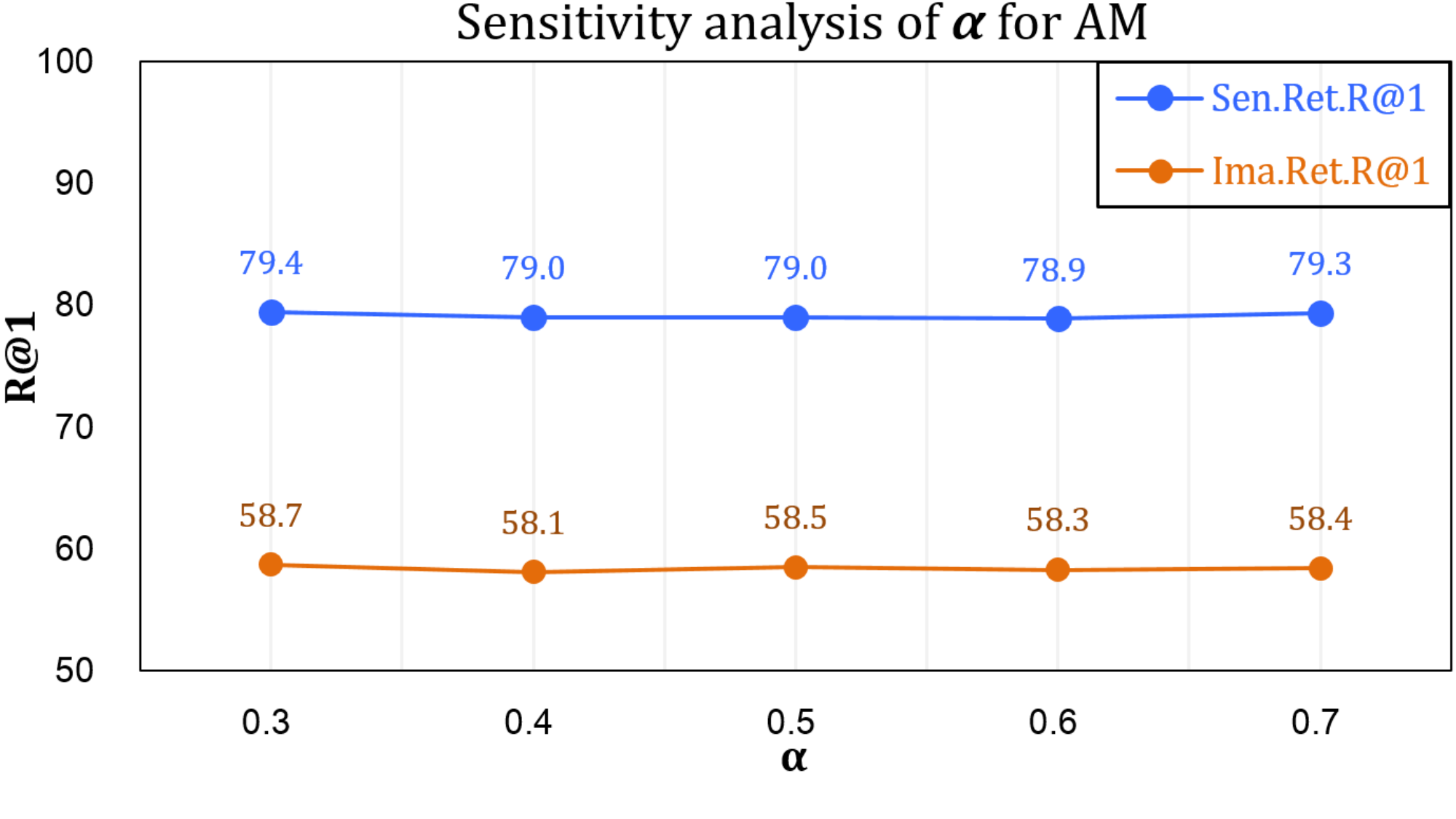}
    \end{tabular}
    \caption{Analyses of hyperparameter $\gamma$ for RM and $\alpha$ for AM under OSS.}
    \label{fig:hyper-parameter}
\end{figure}

\textbf{Comparison with Deep Metric Learning.}
In TABLE~\ref{tab:MetLea}, we report the retrieval results with several popular loss functions based on single branch, which focus on cross-modal projection~\cite{ITM:CMPL}, hard negative mining~\cite{ITM:VSE++,ITM:MPL}, and adaptive margin setting~\cite{ITM:Ladder,ITM:SAM} respectively. 
Here, we utilize our proposed DBL under MSS for comparison, given its relatively minimal additional consumption costs (only extra 11\% memory and 18\% time costs), when compared to the single branch with various loss functions. 
We can find that almost all similarity metrics learned through the above losses are better than the original ranking loss~\cite{ITM:DSPE}, confirming that it is beneficial to highlight the informative triplets and develop the appropriate thresholds. It is obvious that our DBL strategy achieves more impressive improvements and outperforms the best competitor MPL~\cite{ITM:MPL} on bidirectional R@1 by 1.3/1.7\% and 2.3/2.3\% in the relative/absolute manner separately. Besides, the comparison with the most related metrics~\cite{ITM:Ladder,ITM:SAM} validates that explicit margin constraints by the DBL are more effective and applicable for the network to obtain more powerful matching capabilities across modalities.

\begin{figure*}[t!]
    \centering
    \begin{tabular}{@{}c}
    \includegraphics[width=0.98\linewidth, height=0.23\linewidth,trim= 0 330 70 0,clip]{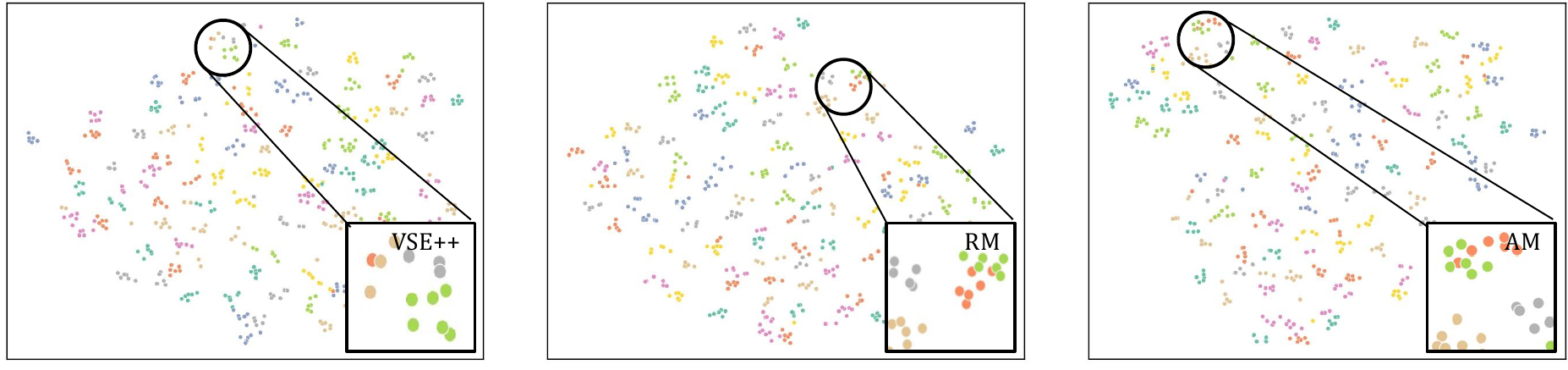}
    \end{tabular}
    \caption{Comparison of feature distribution for image and text samples. We implement t-SNE to visualize the image and text features based on GPO~\cite{ITM:GPO}.}
    \label{fig:feature_visualizaiton}
\end{figure*}

\begin{figure}[t!]
	\centering
	\begin{tabular}{@{}c}
		\includegraphics[width=0.98\linewidth, 
		height=0.5\linewidth,trim= 40 210 360 20,clip]{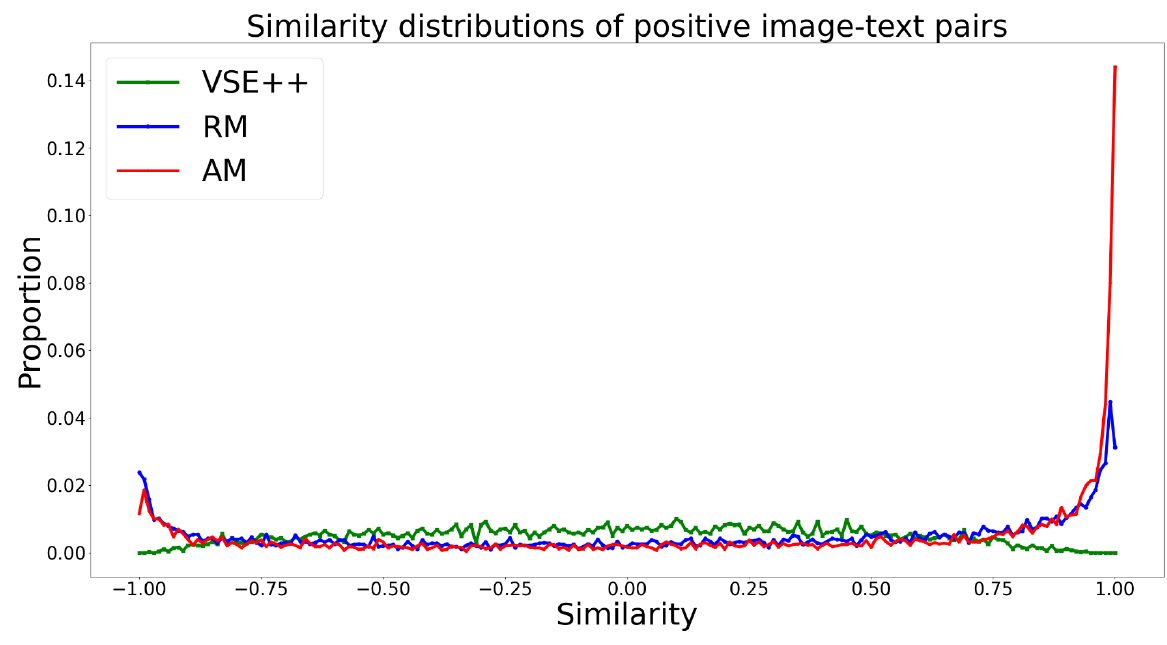}
	\end{tabular}
	\caption{Comparison of similarity distribution for positive image-text pairs.}
	\label{fig:pos_dis}
\end{figure}

\subsection{Ablation Studies}
\label{subsecABS}

\textbf{Configurations of boosting loss.}
TABLE~\ref{tab:Config} shows the different configurations of our DBL, consisting of Sum-Max sampling, Fixed-Soft margin value, and Relative-Absolute boosting manners. 
\textbf{1) Sum vs. Max.} Compared with Model \#B and \#C via Sum operation, Model \#F and \#G show that Max operation can excavate more valuable triplets and obtain the 1.7/0.8\% and 1.0/1.4\% R@1 increase on sentence and image retrieval for the Relative/Absolute strategy.
\textbf{2) Fixed vs. Soft.} Model \#D, \#E, \#F, and \#G demonstrate that the Soft form displays stronger abilities of image retrieval but achieves slightly attractive promotions on sentence retrieval, while the Fixed form obtains the optimal balance between image and sentence retrievals. Note that the Soft strategy stresses more on the hard triplets, making its gradient easily dominated by noise, being a result of either deficiency of model or data per se.
\textbf{3) Relative vs. Absolute.} Comparing Model \#B-\#G, and more applications in TABLE~\ref{tab:cocof30k}, we discover that both of them display the general effectiveness and unique qualities on R@$\kappa$ under diverse network architectures. Based on a more intuitive MD metric, the Absolute manner produces more compact similarity distributions than the Relative one.
\textbf{4) Relative \&. Absolute.} We average the relative and absolute boosting losses to jointly supervise the training process. Compared with Model \#F and \#G, Model \#H and \#I display no further improvements on both R@$\kappa$ and MD metrics. This may be because AM imposes tighter constraints than RM, and such simple combinations fail to reinforce the penalty and further widen the distance between positive and negative pairs.

\textbf{Analyses of hyperparameter tuning.} The exclusive $\alpha$ and $\gamma$ tuning are shown in Fig.~\ref{fig:hyper-parameter}.
For $\alpha=0.3-0.7$, AM brings at least 3.2/1.9$\%$ R@1 gains on sentence and image retrievals. 
For $\gamma=0.1-0.3$, RM obtains a steady R@1 gain of at least 1.9/1.3$\%$ at two directions. 
Note that the VSE++ loss~\cite{ITM:VSE++} produces varied results with different margin $\gamma$, and $\gamma=0.2$ is a commonly-used configuration~\cite{ITM:VSRN,ITM:GPO,ITM:SGRAF,ITM:NAAF}. 
Hence for simplicity and fair comparison, we set $\alpha=0.5$ and $\gamma=0.2$ in all our experiments, and obtain robust and stable performance benefits over various datasets and approaches.

\begin{figure}[t!]
    \centering
    \begin{tabular}{@{}c}
    \includegraphics[width=0.98\linewidth, 
    height=0.5\linewidth,trim= 40 210 360 20,clip]{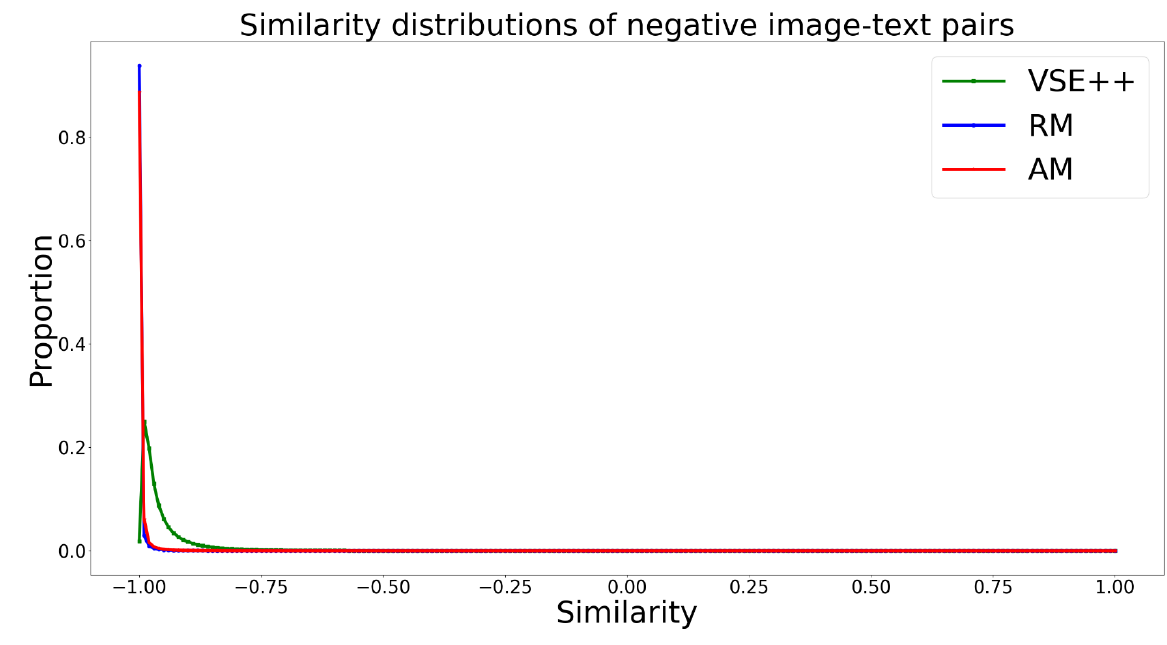}
    \end{tabular}
    \caption{Comparison of similarity distribution for negative image-text pairs.}
    \label{fig:neg_dis}
\end{figure}

\textbf{Impact of the well-trained anchor branch.} To validate this, we take AM under OAS on Flickr30K as an example. \textbf{(1)} We utilize $\gamma$=0.1 of hard ranking loss~\cite{ITM:VSE++} and 50$\%$ Flickr30k training data to construct two kinds of sub-optimal anchor branches. Interestingly, DBL still improves corresponding target branches by 2.7/1.9\% and 7.4/5.8\% R@1 gains. \textbf{(2)} We also use VSE++~\cite{ITM:VSE++} as a poor backbone that outputs many noisy matching results, which achieves R@1 benefits of 1.8/1.3\% by DBL. \textbf{(3)} They verify that even with an inferior reference, DBL still displays good stability and robustness.

\textbf{Feature distributions of image and text samples.}
We visualize the feature distribution of image and text features in Fig.~\ref{fig:feature_visualizaiton}. Note that for cross-modal interaction methods, feature visualization does not provide a direct reflection of the similarity measurement across modalities. Therefore, we adopt GPO~\cite{ITM:GPO} based on mono-modal representation for better illustration. We can observe that our DBL achieves better feature separation, even for some challenging samples. The anchor branch provides explicit and targeted distance information, which enables the target branch to enlarge the gap between image and text through our DBL strategy.

\begin{figure*}[ht]
    \centering
    \begin{tabular}{@{}c}
    \includegraphics[width=0.95\linewidth, height=0.48\linewidth, trim=0 60 220 5,clip]{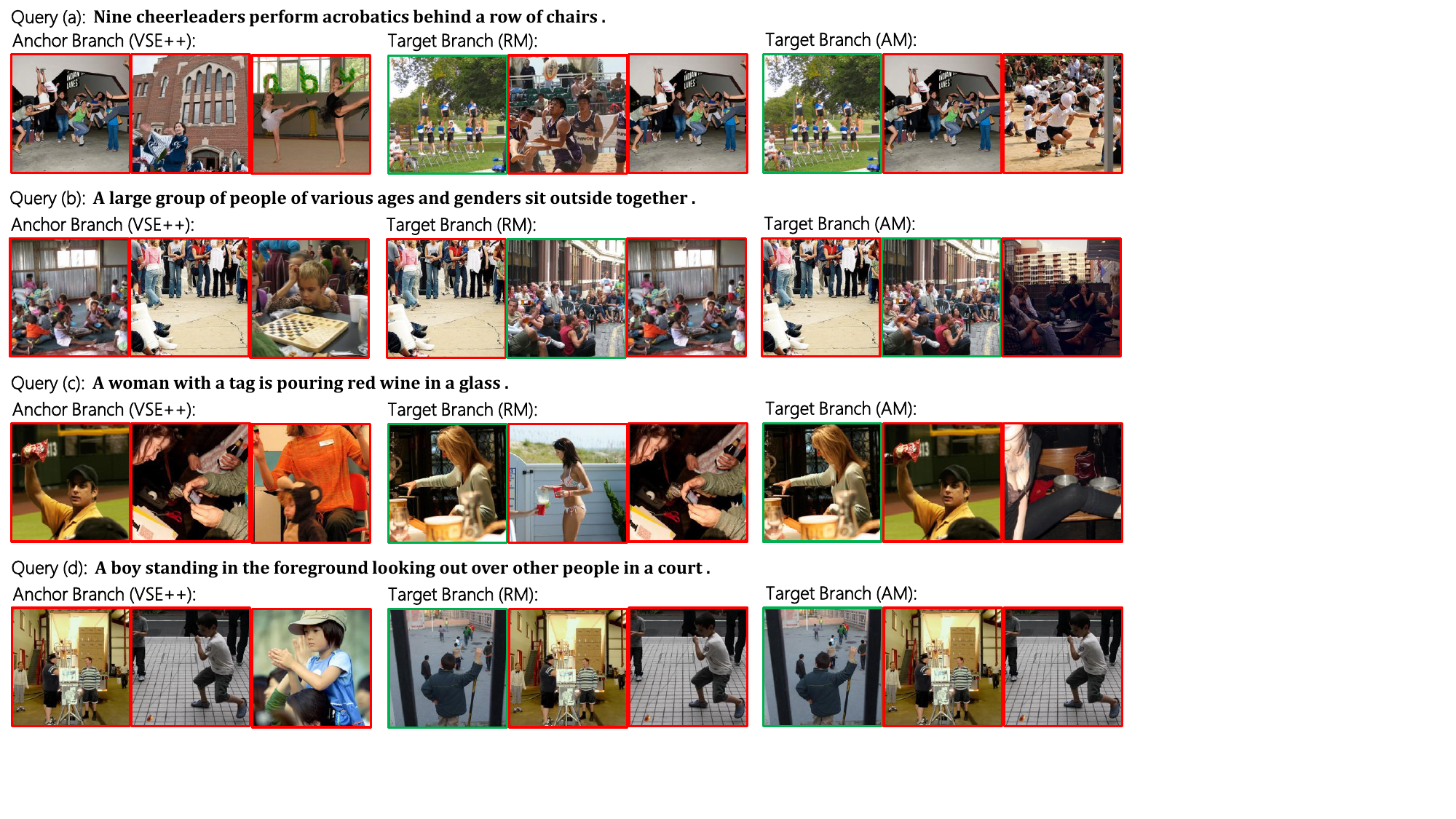}
    \end{tabular}
    \caption{Several retrieval examples on image retrieval. Green denotes the ground-truth image candidates and red denotes the unmatched retrieval samples.}
	\label{fig:t2ipp}
\end{figure*}

\begin{figure*}[ht]
    \centering
    \begin{tabular}{@{}c}
    \includegraphics[width=0.95\linewidth, height=0.48\linewidth, trim=0 0 85 0,clip]{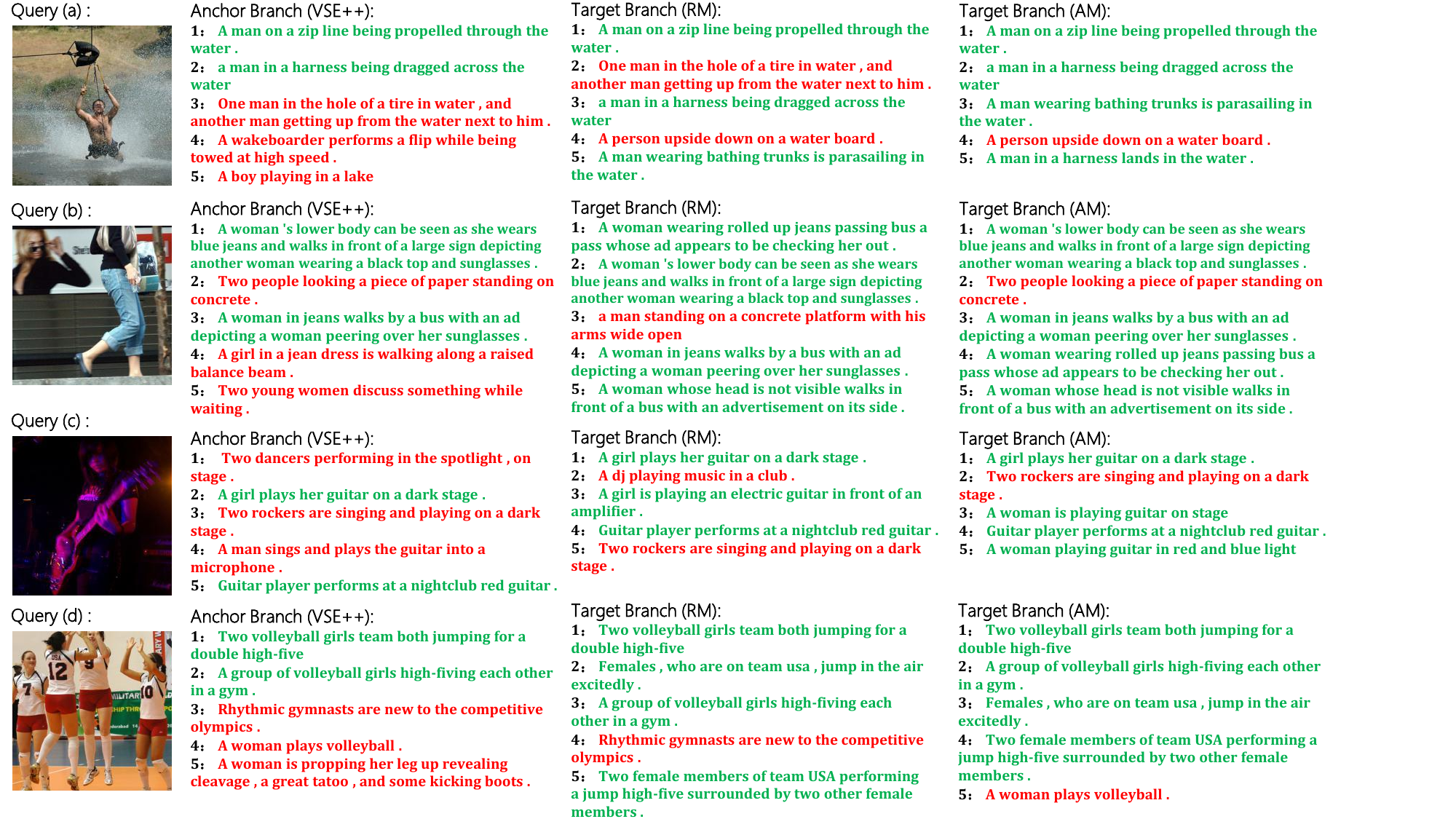}
    \end{tabular}
    \caption{Several retrieval examples on sentence retrieval. Green denotes the ground-truth sentence candidates and red denotes the unmatched retrieval samples.}
    \label{fig:i2tpp}
\end{figure*}

\textbf{Similarity distributions of image-text pairs.}
Fig.~\ref{fig:pos_dis} and~\ref{fig:neg_dis} exhibit the similarity distributions of positive and negative pairs respectively on the Flickr30K test set. With VSE++ loss~\cite{ITM:VSE++}, the scores of negative pairs are concentrated near the value -1, while the curve of positive ones is relatively smooth. After introducing RM, their values are more densely distributed at the value -1 and 1 respectively, and the separability between them is fully exploited. Besides, AM can further enlarge the variations between matched and unmatched pairs, confirming that learning an adaptive and explicit margin can lead to sufficient distance metrics and powerful matching patterns. Note that a slight peak arises around -1.0 for positive pairs. This is because DBL generally produces more confident predictions as compared to plain VSE++, and could inadvertently misclassify some hard positive pairs with weak correlations as negative samples. We further discover that the positive pairs with the similarity between -1.0 and -0.8 by RM and AM are mostly distributed ranging from -0.9 to -0.2 in the original VSE++ with 87.8\% and 92.6\% probability.

\textbf{Qualitative results of image and sentence retrieval.}
Fig.~\ref{fig:t2ipp} and~\ref{fig:i2tpp} display several retrieval examples on image and sentence retrieval, which can qualitatively indicate the learned distance measure between image and text. Compared with the anchor branch, the target branch by our DBL is capable of better recognizing cross-modal contents, and effectively distinguishing the correct results from various distractions with similar semantics, which validates the powerful applicability and matching capability of our DBL strategy.

\section{Conclusion and Future Works}
In this paper, we propose a novel Deep Boosting Learning (DBL) strategy to seek a powerful modeling capability by imposing an adaptive and dynamic boosting mechanism for image-text matching task. Specifically, we first plumb the model property and data representation thoroughly, which in turn facilitates the learning process with appropriate regulations in a boosting manner. Extensive experiments on two benchmark datasets validate that our DBL further enlarges the insufficient variations within triplets and exploits the optimal feature and distance metrics across modalities. Besides, we discover that DBL consistently improves various popular frameworks by a large margin, confirming its general effectiveness and flexible applicability in image-text matching field. As one of the peer-to-peer strategies, DBL goes deeper than many related cooperative methods by learning margin knowledge to gain greater benefits under the same training and inference schemes, from which the community may get inspiration. In the future, we would like to incorporate them simultaneously in self-branch boosting and cross-branch imitating manners.

\ifCLASSOPTIONcaptionsoff
  \newpage
\fi
\bibliographystyle{IEEEtran}
\bibliography{IEEEabrv,refs}
\end{document}